\documentclass[9pt,twocolumn,twoside]{arxiv_pnas-class}

\templatetype{arxivpnas} 

\usepackage{lipsum}  
\usepackage{tikz}
\usetikzlibrary{positioning, arrows.meta, decorations.markings}
\usepackage{nicefrac}
\usepackage[capitalize,noabbrev]{cleveref}
\usepackage{hyperref}
\def\x{{\mathbf{x}}}

\def\O{\mathcal{O}}
\def\o{o}
\newcommand{\beq}{\begin{equation}}
\newcommand{\eeq}{\end{equation}}
\newcommand{\beqs}{\begin{equation*}}
\newcommand{\eeqs}{\end{equation*}}
\newcommand{\beqn}{\begin{eqnarray}}
\newcommand{\eeqn}{\end{eqnarray}}
\newcommand{\lpa}{\left(}
\newcommand{\rpa}{\right)}
\newcommand{\lsq}{\left[}
\newcommand{\rsq}{\right]}

\newcommand{\createtree}[5]{
  \pgfmathtruncatemacro{\nextlayer}{#1+1}
  \ifnum#1=#2 
    \node [varnode] (#3) [below left= of #4] {};
  \else
      \node [factnode] (#3) [below= of #4] {};
      \draw (#4) -- (#3);
    \foreach \x in {1,2}{
      \pgfmathtruncatemacro{\newname}{#3*10+\x} 
      \node [varnode] (\newname) [below= of #3, xshift=(2*\x-3)*(40/#1)] {$X_{\x}$};
      \draw (#3) -- (\newname);
      }

  \fi
}

\newcommand{\nnu}[1]{{\nu_{\uparrow}^{(#1)}}}
\newcommand{\ncu}[1]{{\tilde{\nu}_{\uparrow}^{(#1)}}}
\newcommand{\nnd}[1]{{\nu_{\downarrow}^{(#1)}}}
\newcommand{\ncd}[1]{{\tilde{\nu}_{\downarrow}^{(#1)}}}
\newcommand{\rul}[1]{{\rr^{(#1)}}}
\newcommand{\rr}{{\psi}}

\newcommand{\corr}[1]{{\overline{#1}}}
\newcommand{\bel}{\mathcal{B}}
\newcommand{\samp}{\mathcal{S}}
\newcommand{\ham}[1]{\Delta_{#1}}
\newcommand{\asso}{\{y\}\leftarrow\{\x\}}

\definecolor{C0}{HTML}{1F77B4}
\definecolor{C1}{HTML}{FF7F0E}
\definecolor{C2}{HTML}{2CA02C}
\definecolor{C3}{HTML}{D62728}
\definecolor{C4}{HTML}{9467BD}
\definecolor{C5}{HTML}{8C564B}
\definecolor{C6}{HTML}{E377C2}
\definecolor{C7}{HTML}{7F7F7F}
\definecolor{C8}{HTML}{BCBD22}
\definecolor{C9}{HTML}{17BECF}
\definecolor{Gold}{HTML}{FFD700}
\definecolor{Navy}{HTML}{000080}

\def\appBayes{\protect\hyperlink{app:bayes}{SI Appendix 1}}
\def\appBP{\protect\hyperlink{app:bp}{SI Appendix 2}}
\def\appUpAnneal{\protect\hyperlink{app:anneal}{SI Appendix 2.A}}
\def\appMapping{\protect\hyperlink{app:CC}{SI Appendix 3}}
\def\appResnet{\protect\hyperlink{app:resnets}{SI Appendix 4}}
\def\appGauss{\protect\hyperlink{app:gaussian-mixture}{SI Appendix 5}}

\def\secImageNetCNN{\protect\hyperlink{sec:imagenet_representations}{Section 1.C}}
\def\secRHM{\protect\hyperlink{sec:rhm}{Section 2}}
\def\secBP{\protect\hyperlink{sec:bp}{Section 3}}
\def\secMF{\protect\hyperlink{sec:mean-field}{Section 4}}

\def\eqBayes{\autoref{eq:bayes}}
\def\eqBeliefMain{\autoref{eq:belief-main}}

\def\figUDTheory{\autoref{fig:activations_RHM}-(a) }
\def\figUDAct{\autoref{fig:activations_RHM}-(b)}
\def\figImageNet{\autoref{fig:imagenet-main} }
\def\figImageNetAct{\autoref{fig:imagenet-main}-(right) }

\begin{document}

\title{A Phase Transition in Diffusion Models Reveals the Hierarchical Nature of  Data}

\author[a,1]{Antonio Sclocchi}
\author[a,b]{Alessandro Favero}
\author[a,1]{Matthieu Wyart}

\affil[a]{Institute of Physics, Ecole Polytechnique F\'ed\' erale de Lausanne, 1015 Lausanne, Switzerland}
\affil[b]{Institute of Electrical and Micro Engineering, Ecole Polytechnique F\'ed\' erale de Lausanne, 1015 Lausanne, Switzerland}

\leadauthor{Sclocchi}

\significancestatement{
The success of deep learning
is often attributed to its ability to harness the hierarchical and compositional structure of data. However, formalizing and testing this notion remained a challenge. This work shows how diffusion models -- generative AI techniques producing high-resolution images -- operate at different hierarchical levels of features over different time scales of the diffusion process. This phenomenon allows for the generation of images of various classes by recombining low-level features. We study a hierarchical model of data that reproduces this phenomenology and provides a theoretical explanation for this compositional behavior.
Overall, the present  framework provides a description of how generative models operate, and put forward diffusion models as powerful lenses to probe data structure.
}

\correspondingauthor{\textsuperscript{1}To whom correspondence should be addressed. E-mail: antonio.sclocchi@epfl.ch; matthieu.wyart@epfl.ch.}


\begin{abstract}
Understanding the structure of real data is paramount in advancing modern deep-learning methodologies. Natural data such as images are believed to be composed of features organized in a hierarchical and combinatorial manner, which neural networks capture during learning. Recent advancements show that diffusion models can generate high-quality images, hinting at their ability to capture this underlying compositional structure. We study this phenomenon in a hierarchical generative model of data. We find that the backward diffusion process acting after a time $t$ is governed by a phase transition at some threshold time, where the probability of reconstructing high-level features, like the class of an image, suddenly drops. Instead, the reconstruction of low-level features, such as specific details of an image, evolves smoothly across the whole diffusion process. This result implies that at times beyond the transition, the class has changed, but the generated sample may still be composed of low-level elements of the initial image. We validate these theoretical insights through numerical experiments on class-unconditional ImageNet diffusion models. Our analysis characterizes the relationship between time and scale in diffusion models and puts forward generative models as powerful tools to model combinatorial data properties.
\end{abstract}

\doi{\url{www.pnas.org/cgi/doi/10.1073/pnas.XXXXXXXXXX}}

\maketitle
\thispagestyle{firststyle}
\ifthenelse{\boolean{shortarticle}}{\ifthenelse{\boolean{singlecolumn}}{\abscontentformatted}{\abscontent}}{}

\firstpage[4]{4}

\dropcap{U}nderstanding which data are learnable by algorithms is key to machine learning. Techniques such as supervised, unsupervised, or self-supervised learning are most often used with high-dimensional data. However, in large dimensions, for generic data or tasks, learning should require a number of training examples that is exponential in the dimension \cite{luxburg2004distance,bach2017breaking}, which is never achievable in practice. 
The success of these methods with limited training set sizes implies that high-dimensional data such as images or text are highly structured.
In particular, these data are believed to be composed of features organized in a hierarchical and compositional manner \cite{patel2015probabilistic, mossel2016deep, poggio2017why, malach2018provably, schmidt2020nonparametric, favero_locality_2021, cagnetta2023can, cagnetta2023deep}. Arguably, generative models can compose a whole new datum by assembling features learned from examples.
Yet, formalizing and testing this idea is an open challenge.
In this work, we show how diffusion models \cite{sohl2015deep,ho2020denoising,song2019generative, song2020score} -- such as DALL$\cdot$E \cite{betker2023improving} and StableDiffusion  \cite{rombach2022high} --
generate images by composing features at different hierarchical levels throughout the diffusion process
. Specifically, we first provide quantitative evidence of compositional effects in the denoising diffusion of images. We then provide a theoretical characterization of such effects through a synthetic model of hierarchical and compositional data.

\begin{figure}[H]
    \centering
    \includegraphics[width=\columnwidth]{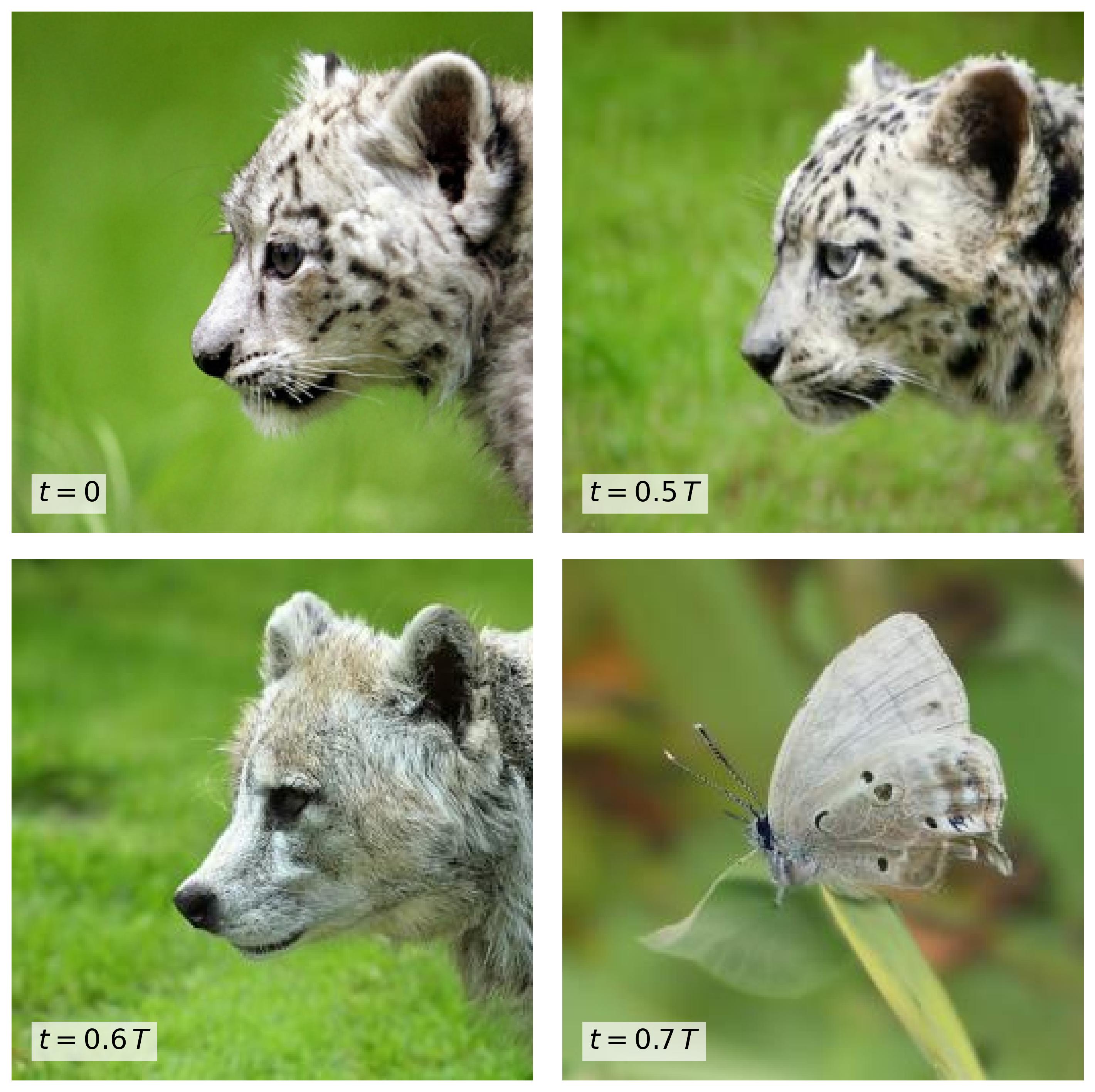}
    \vspace{-10pt}
    \caption{\textbf{Illustration of forward-backward experiments.} Images generated by a denoising diffusion probabilistic model starting from the top-left image and inverting the dynamics at different times $t$. $T$ corresponds to the time scale when the forward diffusion process converges to an isotropic Gaussian distribution. At small $t$, the class of the generated image remains unchanged, with only alterations of low-level features, such as the eyes of the leopard. After a characteristic time $t$, the class undergoes a phase transition and changes. However, some low-level attributes of the original image are retained to compose the new image. For instance, the wolf is composed of eyes, nose, and ears similar to those of the leopard, and the butterfly inherits its colors and black spots.}
    \label{fig:splash-figure}
\end{figure}

Diffusion models add noise to images as time increases and learn the reverse denoising process to generate new samples. 
In particular, if some finite amount of noise is added to an image and the process is then reversed, we observe that: 
\textit{(i)} for small noise, only low-level features of the image change
; \textit{(ii)} at a threshold noise, the probability of remaining in the same class suddenly drops to near-random chance; \textit{(iii)}  beyond that point, low-level features may persist and compose the element of a new class. 
While observation \textit{(i)} is intuitive and was first noticed in \cite{ho2020denoising}, the fact that at large noise only low-level features may remain unchanged is surprising. We will show below that this property is expected for hierarchical data.
These results appear already evident in examples such as \autoref{fig:splash-figure}, and we systematically quantify them by considering the change of internal representations in state-of-the-art convolutional neural networks.

We show that the observations \textit{(i)}, \textit{(ii)}, and \textit{(iii)} can be theoretically explained through generative models of data 
with a hierarchical and compositional structure \cite{cagnetta2023deep}, inspired by models of formal grammars and statistical physics. We demonstrate that for these models the Bayes optimal denoising can be described exactly using belief propagation on tree-like graphs. Remarkably, our analysis predicts and explains both the phase transition in the class (observation \textit{(ii)}) and how lower-level features compose to generate new data before and after this transition (observations \textit{(i)} and \textit{(iii)}).

Overall, our results reveal that diffusion models act at different hierarchical levels of the data at different time scales within the diffusion process
and establish hierarchical generative models as valuable theoretical tools to address several unanswered questions in machine learning.

\subsection{Our contributions}

We perform a systematic study of the denoising diffusion dynamics on ImageNet. 
We invert the noising process at some time $t$, leading to novel noiseless images. We then analyze how the representation of state-of-the-art convolutional architectures changes between the initial and newly generated images as a function of both time $t$ and depth of the representation. This analysis reveals the presence of a sharp transition in the class at a given time or noise level. Importantly, at times beyond the transition, when the class has changed, we find that the generated images may still be composed of low-level features of the original image.

\looseness=-1 To model theoretically the compositional structure of images, we consider hierarchical generative models of data where the structure of the latent variables is tree-like. We use belief propagation to study the optimal denoising dynamics for such data and obtain the evolution of latent variables' probabilities for different levels of corruption noise. 
In the limit of a large tree depth, this analysis reveals a phase transition for the probability of reconstructing the root node of the tree -- which represents the class label of a data point -- at a specific noise threshold. Conversely, the probability of reconstructing low-level latent variables evolves smoothly throughout the denoising diffusion process. Thus, after the transition, low-level features of the original datum may persist in composing a generated element of a new class, as we empirically observe in ImageNet.  Finally, we show numerically that the dynamics of the latent variables is reflected in the hidden representation of deep networks previously trained on a supervised classification task on these data.

\paragraph{Organization of the paper}
In \Cref{sec:diffusion}, we introduce denoising diffusion probabilistic
models and present our large-scale experiments on ImageNet data.
In \Cref{sec:rhm}, we define the hierarchical generative model of data that we study theoretically.
In \Cref{sec:bp}, we study the optimal denoising for these data using message-passing techniques and show that our model captures the experimental observations on real data.
In \Cref{sec:mean-field}, we perform a mean-field analysis of the optimal denoising process, obtaining an analytical prediction for the phase transition of the class at a critical noise value and for the reconstruction probabilities of lower-level features.

\subsection{Related work}

\paragraph{Forward-backward protocol in diffusion-based models} 

\cite{ho2020denoising} introduced the ``forward-backward" protocol to probe diffusion-based models, whereby an image with a controlled level of noise is then denoised using a reverse-time diffusion process. It led to the observation that ``when the noise is small, all but fine details are preserved, and when it is large, only large scale features are preserved''. Although our work agrees with the first part of the statement, it disagrees with the second.
Our work also provides a systematic quantification of the effects of forward-backward experiments, going beyond
qualitative observations based on individual images as in \cite{ho2020denoising}. Specifically, we introduce quantitative observables that characterize changes in the latent features of images and perform extensive experiments with state-of-the-art models, averaging results over $10^5$ ImageNet samples. Such quantification is key to connecting with theory.
The forward-backward protocol was also studied in \cite{behjoo2023u} to speed up the generation process of images.

\paragraph{Theory of diffusion models}

Most of the theoretical work on diffusion models considers simple models of data. Under mild assumptions on the data distribution, diffusion models exhibit a sample complexity that scales exponentially with the data dimension \cite{block2020generative,oko2023diffusion}. This curse of dimensionality can be mitigated through stronger distributional assumptions, such as considering data lying within a low-dimensional latent subspace \cite{de2022convergence,chen2023score,yuan2023reward},  Gaussian mixture models \cite{biroli2023generative,shah2023learning, Cui2023AnalysisOL}, graphical models \cite{mei2023deep}, or data distributions that can be factorized across scales \cite{kadkhodaie2023learning}. 
For multimodal distributions such as Gaussian mixtures, the backward dynamics exhibits a cross-over time when it concentrates toward one of the modes  \cite{biroli2023generative, ambrogioni2023statistical, raya2024spontaneous}. This cross-over is similar to our observation (ii) above if these modes are interpreted as classes.
As demonstrated in {\appGauss}, such models of data cannot reproduce our salient predictions and observations.
Closer to our work, \cite{okawa2023compositional} considers synthetic compositional data to empirically show how diffusion models learn to generalize by composing different concepts. In contrast, we study data that are not only compositional but also hierarchically structured and make quantitative predictions on how diffusion models compose features at different scales.  

\paragraph{Hierarchical models of natural data}
Generative models of data have a long history of describing the structure of language and image data.
In linguistics, formal grammars describe the syntactic structure of a language through a hierarchical tree graph \cite{rozenberg_handbook_1997}.
Similar ideas have been explored to decompose visual scenes hierarchically into objects, parts, and primitives \cite{zhu2007stochastic} and have been formalized in pattern theory \cite{stoyan1997grenander}. These hierarchical models led to practical algorithms for semantic segmentation and scene understanding, as illustrated in, e.g., \cite{jin2006context, siskind2007spatial, li2009towards}. Recent works propose a hierarchical decomposition of images, in which latent variables are wavelet coefficients at different scales 
\cite{marchand2022wavelet,
kadkhodaie2023learning}. In this case, the graph is not tree-like \cite{
kadkhodaie2023learning} -- a conclusion that could stem from the specific choice of latent variables.

\paragraph{Hierachical models in machine learning theory}

More recently, generative models of data received attention in the context of machine learning theory.
In supervised learning, deep networks can represent hierarchical tasks more efficiently than shallow networks~\cite{poggio2017why} and can efficiently learn them from an information theory viewpoint~\cite{schmidt2020nonparametric}.
For hierarchical models of data, correlations between the input data and the task are critical for learning~\cite{mossel2016deep,shalev2017failures,malach2018provably,malach2020implications} and the representations learned by neural networks with gradient descent reflect the hidden latent variables of such models both in Convolutional Neural Networks (CNNs) \cite{cagnetta2023deep} and transformers \cite{allen2023physics}. In this work, we use these hierarchical generative models of data to study the denoising dynamics of diffusion models theoretically.

\section{Diffusion models and feature hierarchies}\label{sec:diffusion}

\begin{figure*}
    \centering
    \begin{tikzpicture}
    \node[anchor=north west,inner sep=0pt] at (0,0){
    \resizebox{.95\columnwidth}{!}{\includegraphics[width=1\columnwidth]{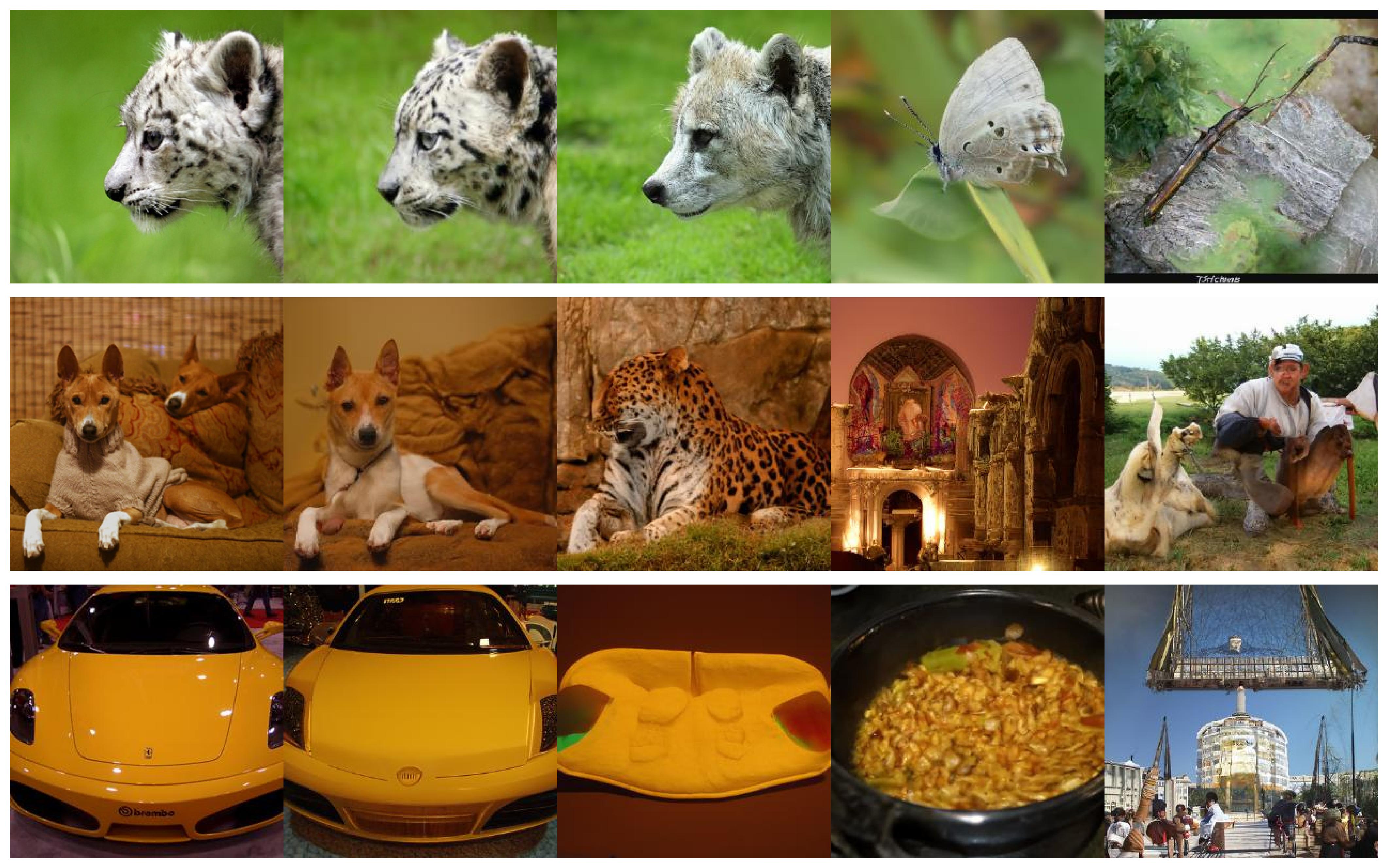}}};
    \node[font=\small] at (6ex,-38.5ex) {${t=0}$};
    \node[font=\small] at (18ex,-38.5ex) {${t=0.5\,T}$};
    \node[font=\small] at (30ex,-38.5ex) {${t=0.6\,T}$};
    \node[font=\small] at (41ex,-38.5ex) {${t=0.7\,T}$};
    \node[font=\small] at (53ex,-38.5ex) {${t=T}$};
    \end{tikzpicture}
    \hspace{.1cm}
    \includegraphics[width=1.\columnwidth]{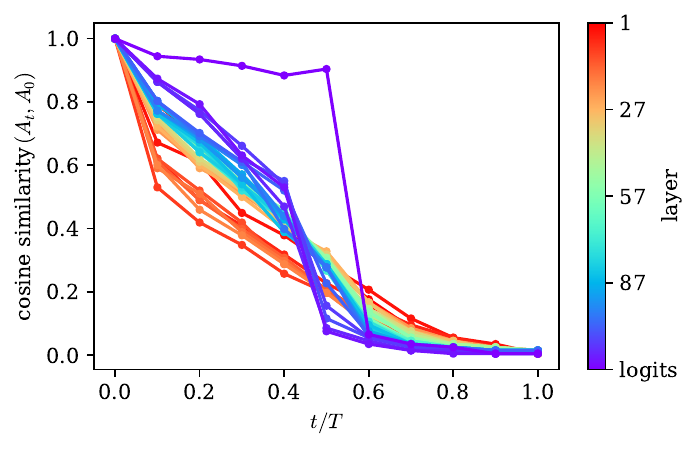}

    \caption{\textit{Left panel.} \textbf{Examples of images generated by reverting the diffusion process at different times $t$.} Starting from the left images $x_0$ at time $t=0$, we generate samples $\hat{x}_{0}(t)\sim  p_\theta(\hat{x}_{0}|x_t)$ by first running the diffusion process up to time $t$ and then reverting it, as described in \Cref{sec:forward-backward-exp}.
    At time $t=T$, $x_T$ corresponds to isotropic Gaussian noise and the generated image $\hat{x}_{0}(T)$ is uncorrelated from $x_0$. At intermediate times, instead, a sudden change of the image class is observed, while some lower-level features are retained.
    \textit{Right panel.} \textbf{Cosine similarity between the post-activations of the hidden layers of a ConvNeXt Base \cite{liu2022convnet} for the initial images $x_0$ and the synthesized ones $\hat{x}_{0}(t)$.} Around $t \approx T/2$, the similarity between logits exhibits a sharp drop, indicating the change in class, while the hidden representations of the first layers change more smoothly. This indicates that certain low-level features from the original images are retained for composing the sampled images also after the class transition. To compute the cosine similarity, all activations are standardized, i.e., centered around the mean and scaled by the standard deviation computed on the 50000 images of the ImageNet-1k validation set. At each time, the values of the cosine similarity correspond to the maximum of their empirical distribution over $10000$ images ($10$ per class of ImageNet-1k).}
    \label{fig:imagenet-main}
\end{figure*}

This section introduces denoising diffusion probabilistic models and demonstrates how class-unconditional ImageNet diffusion models operate on image features across different hierarchical levels at different time scales.\footnote{The code for running the experiments on ImageNet is available at \href{https://github.com/pcsl-epfl/forward-backward-diffusion}{github.com/pcsl-epfl/forward-backward-diffusion} \cite{sclocchiGitHub}.}

\subsection{Background on denoising diffusion models}

Denoising diffusion probabilistic models (DDPMs) \cite{ho2020denoising} are generative models designed to sample from a distribution by reversing a step-by-step noise addition process. In particular, let $q(\cdot)$ represent the data distribution, and let $x_0$ be a sample drawn from this distribution, i.e., $x_0 \sim q(x_0)$. First, DDPMs consist of a \textit{forward process} which is a Markov chain generating a sequence of noised data \{$x_t\}_{1\leq t \leq T}$ by introducing isotropic Gaussian noise at each time step $t$ with a variance schedule \{$\beta_t\}_{1\leq t \leq T}$ as follows:
\begin{align}
     q(x_1,\dots,x_T|x_0) &= \prod_{t=1}^T q(x_t|x_{t-1}) \nonumber \\  
     &= \prod_{t=1}^T \mathcal{N}(x_t;\sqrt{1-\beta_t}x_{t-1},\beta_t \mathbb{I}).
\end{align}
Thus, at each time step $t$, we have
\begin{equation}\label{eq:diffused-xt}
    x_t = \sqrt{\overline{\alpha_t}} x_0 + \sqrt{1-\overline{\alpha_t}} \eta
\end{equation}
with $\overline{\alpha_t} = \prod_{t'=1}^t (1-\beta_{t'})$ and $\eta \sim \mathcal{N}(0,\mathbb{I})$. By selecting the noise schedule such that $\overline{\alpha}_t \to 0$ as $t \to T$, the distribution of $x_T$ becomes an isotropic Gaussian distribution. Subsequently, DDPMs reverse this process by gradually removing noise in a \textit{backward process}. In this process, the models learn Gaussian transition kernels $q(x_{t-1}|x_t)$ by parametrizing their mean and variance using a neural network with parameters $\theta$ as follows:
\begin{equation}\label{eq:ddpm-network}
    p_\theta(x_{t-1}|x_t) = \mathcal{N}(x_{t-1};\mu_\theta(x_t,t),\Sigma_\theta(x_t,t)).
\end{equation}
After training, the learned $p_\theta$ can be used to generate novel examples by initiating the process with $x_T \sim \mathcal{N}(0,\mathbb{I})$ and running it in reverse to obtain a sample from $q$. We refer the reader to \cite{ho2020denoising, nichol2021improved, dhariwal2021diffusion} for more details regarding the formulation of DDPMs and the technical aspects of the reverse transition kernels parameterization with neural networks.

\subsection{Forward-backward experiments}\label{sec:forward-backward-exp}

Previous studies on DDPMs noted that inverting the diffusion process at different times $t$ starting from an image $x_0$ results in samples $\hat{x}_{0}(t) \sim p_\theta(\hat{x}_0|x_t)$ with distinct characteristics depending on the choice of $t$. Specifically, when conditioning on the noisy samples $x_t$'s obtained by diffusing images from the CelebA dataset, one finds that for small values of $t$, only fine details change  \cite{ho2020denoising}.
We conduct a similar experiment using a class-unconditional DDPM introduced by \cite{dhariwal2021diffusion}, on the ImageNet dataset with 256x256 resolution. 

In the left panel of \autoref{fig:imagenet-main}, we present some images resulting from this experiment. For each row, the initial image $x_0$ is followed by images generated by initiating the diffusion process from $x_0$, running the forward dynamics until time $t$, with ${0 < t \leq T=1000}$, and ultimately running the backward dynamics to produce a sample image $\hat{x}_0(t)$.
Our observations from these synthetic images are as follows:

\begin{enumerate}[label = \textit{(\roman*)}]
    \item Similarly to the findings in \cite{ho2020denoising}, at small inversion times $t$, only local features change. Furthermore, the class of the sampled images remains consistent with that of the corresponding starting images, i.e., ${{\rm class}(\hat{x}_0(t))=\rm class}(x_0)$ with high probability.
    \item There exists a characteristic time scale $t^*$ at which the class of the sampled images undergoes a sudden transition.
    \item Even after the class transitions, some low-level features composing the images persist and are reincorporated into the newly generated image. For instance, looking at the left panel of \autoref{fig:imagenet-main}, in the second row, the jaguar is composed with the paws and the ears of the dog in the starting picture, or in the third row, the sofa's armrests inherit the shape of the car headlights.
\end{enumerate}

Our theory, presented in Section \ref{sec:bp} and \ref{sec:mean-field}, predicts how features at different hierarchical levels vary at different time scales of the diffusion dynamics in accordance with observations \textit{(i)}, \textit{(ii)}, and \textit{(iii)}.

\hypertarget{sec:imagenet_representations}{\subsection{ImageNet hidden representations}}
\label{sec:imagenet_representations}

To quantify the qualitative observations mentioned earlier, we design an experiment using the empirically known fact that deep learning models learn hierarchical representations of the data, with complexity increasing as the architecture's depth grows. This phenomenon holds true in both real \cite{olah2020zoom, lecun15, zeiler_visualizing_2014} and synthetic scenarios \cite{cagnetta2023deep, allen-zhu2023how}.
Therefore, we use these internal representations as a proxy for the compositional structure of the data.
We investigate how the hidden representations of a deep ConvNeXt Base model \cite{liu2022convnet}, achieving 96.9\% top-5 accuracy on ImageNet, 
change as a function of the inversion time $t$ and depth $\ell$ of the representation. In the right panel of \autoref{fig:imagenet-main}, we illustrate the value of the cosine similarity between the post-activations of every hidden layer of the ConvNeXt for the initial and generated images. We observe that:
\begin{enumerate}[label = \textit{(\roman*)}]
    \item The representations of early layers of the network, corresponding to low-level and localized features of the images, are the first to change at short diffusion times and evolve smoothly. 
    \item At a specific time and noise scale, the similarity between logits experiences a sharp drop, indicating a transition in the class.
    \item Around the class transition, there is an inversion of the similarity curves. Indeed, the hidden representations in the first layers for the new and generated images now display the largest alignment. This indicates that low-level features from the original images can be reused in composing the sampled images, as qualitatively observed in \autoref{fig:imagenet-main}.
\end{enumerate}
To study the robustness of our results with respect to the architecture choice, in {\appResnet}, we report the same measurements using ResNet architectures with varying width and depth \cite{he_deep_2016}. We find the same qualitative behavior as the ConvNeXt in \Cref{fig:imagenet-main}.

We now present our theory, which predicts these observations. 

\hypertarget{sec:rhm}{\section{Hierarchical generative model of data}}

\label{sec:rhm}

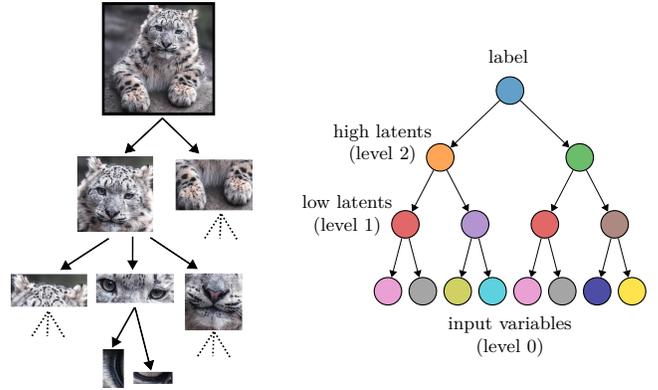
\begin{figure}[t]
    \begin{tikzpicture}
    \node[anchor=north west,inner sep=0pt] at (0,0){\resizebox{0.175\textwidth}{!}{\input{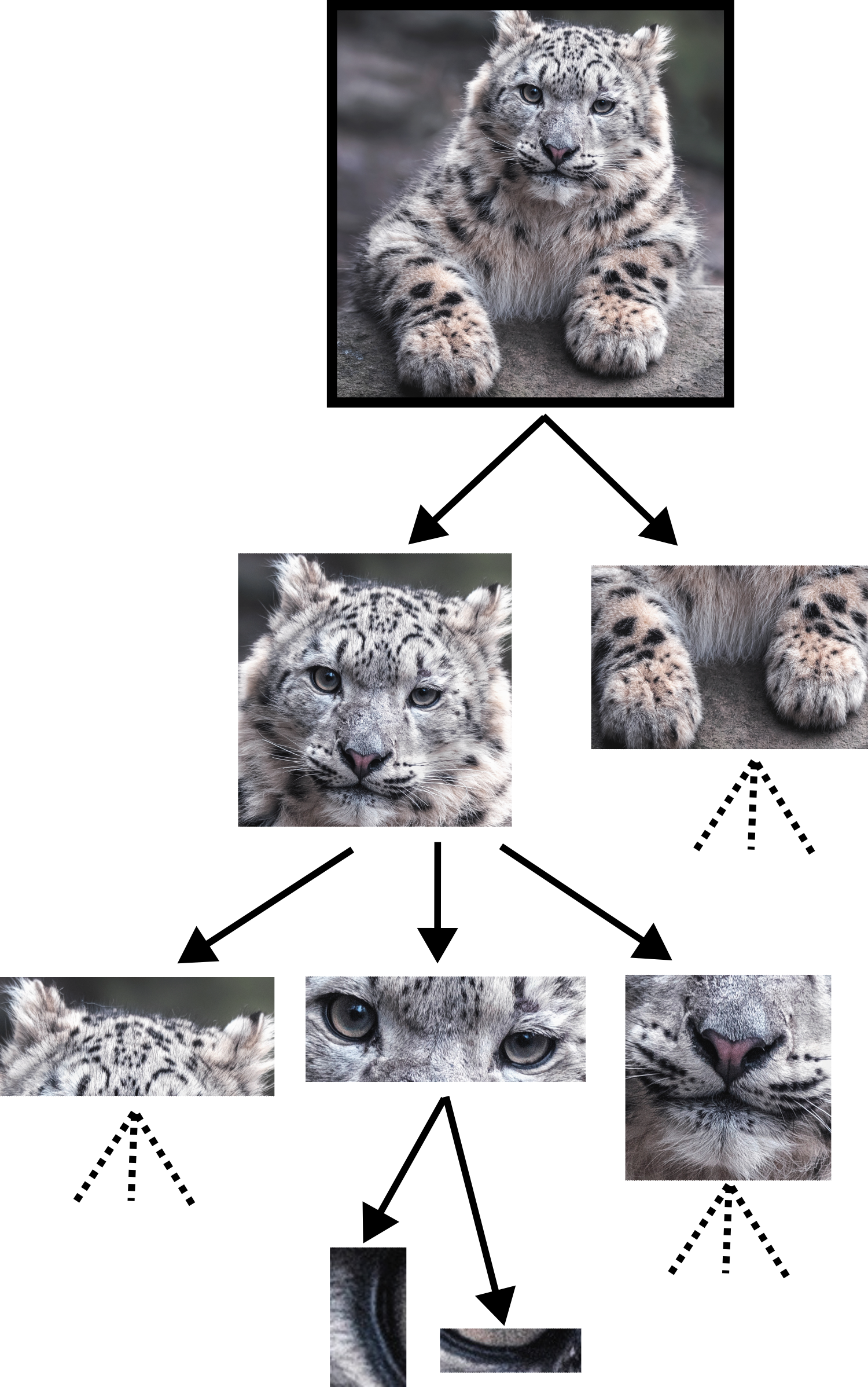}}};
    \end{tikzpicture}
    \begin{tikzpicture}
    \useasboundingbox (-.5,-4.5) rectangle (0,0);
    \node[anchor=north west,inner sep=0pt] at (0,0){\resizebox{0.25\textwidth}{!}{\begin{tikzpicture}[varnode/.style={circle, draw=black, thick, minimum size=7mm, inner sep=0pt},
  factnode/.style={rectangle, draw=black, thick, minimum size=7mm, inner sep=0pt},
  level/.style={sibling distance=100mm/#1}, 
  edge from parent/.style={draw,thick}
  ]

\newcommand{\getcolor}[2]{%
    \ifcase#1
    \ifcase#2 C6\or C7\or C8\or C9\or C6\or C7\or Navy\or Gold\else C9\fi 
    \or
    \ifcase#2 C3\or C4\or C3\or C5\else C3\fi 
    \or
    \ifcase#2 C1\or C2\else C9\fi 
    \else
    C0
    \fi
}

\tikzset{
    varnode/.style={circle, draw, thick, minimum size=7mm, inner sep=0pt, 
    fill=#1,
    fill opacity=0.7},
    factnode/.style={rectangle, draw, thick, minimum size=7mm, inner sep=0pt},
    midarrow/.style={decoration={
            markings,
            mark=at position 0.67 with {\arrow{Stealth}}
        },
        postaction={decorate},
        thick
    },
    connect/.style={decoration={
            markings,
            mark=at position 1.0 with {\arrow{Latex[width=7, length=7]}}
        },
        postaction={decorate},
        thick, line width=1pt},
    MyTip/.tip={Triangle[angle=60:2mm]}
}

\newcommand\layers{3};
\newcommand\initialdistance{.25cm};
\newcommand\laySpace{.06cm};
\foreach \layer in {\layers,...,0}
{
\pgfmathsetmacro\offX{\initialdistance/(2^(\layer+1))};
\pgfmathsetmacro\Xdistance{\initialdistance/(2^\layer)};
\pgfmathsetmacro\numNodes{2^\layer-1};
\foreach \n in {0,...,\numNodes}
{
    \pgfmathsetmacro\Xposition{\offX + \Xdistance * \n};
    \pgfmathsetmacro\Yposition{-\laySpace * \layer};
    \pgfmathtruncatemacro\ColorIndex{mod(\n, 2^\layer)}
    \pgfmathtruncatemacro\LayerIndex{\layers-\layer}
    \node [varnode=\getcolor{\LayerIndex}{\n}] (\layer-\n) at (\Xposition, \Yposition) {};
}
}

\foreach \layer in {\layers,...,1}
{
\pgfmathsetmacro\numNodes{2^\layer-1};
\foreach \n in {0,...,\numNodes}
{
    \pgfmathtruncatemacro\prevLay{\layer-1};
    \pgfmathtruncatemacro\uppNode{int(\n/2)};
    \draw[->, >=MyTip, line width=0.25mm] (\prevLay-\uppNode) -- (\layer-\n);
}
}

\pgfmathsetmacro\offX{\initialdistance/(2^(\layers+1))};
\pgfmathsetmacro\Xdistance{\initialdistance/(2^\layers)};
\pgfmathsetmacro\Xposition{\Xdistance * 2^\layers/2 - 3*\offX};
\pgfmathsetmacro\Yposition{-\laySpace * \layers - 2*\offX};
\node[scale=1.5, align=center] (last_label) at (\Xposition, \Yposition) {input variables\\ 
(level $0$)};

\pgfmathsetmacro\level{\layers-1};
\pgfmathsetmacro\Xposition{-1.5};
\pgfmathsetmacro\Yposition{-\laySpace * \level + 0.5};
\node[scale=1.5, align=center] (last_label) at (\Xposition, \Yposition) {low latents\\ 
(level $1$)};

\pgfmathsetmacro\level{\layers-2};
\pgfmathsetmacro\Xposition{-0.7};
\pgfmathsetmacro\Yposition{-\laySpace * \level + 0.6};
\node[scale=1.5, align=center] (last_label) at (\Xposition, \Yposition) {high latents\\ 
(level $2$)};

\pgfmathsetmacro\level{0};
\pgfmathsetmacro\Xposition{\Xdistance * 2^\layers/2 -.3};
\pgfmathsetmacro\Yposition{-\laySpace * \level + 0.8};
\node[scale=1.5, align=center] (last_label) at (\Xposition, \Yposition) {label};

\end{tikzpicture}}};
    \end{tikzpicture} 
    \caption{\textbf{Sketch of the hierarchical and compositional structure of data.}
    \textit{Left panel:} The leopard in the image can be iteratively decomposed in features at different levels of abstraction.
    \textit{Right panel:} Generative hierarchical model we study in this paper. In this example, depth $L=3$ and branching factor $s=2$. Different values of the input and latent variables are represented with different colors.}
    \label{fig:rhm}
\end{figure}

In this section, we introduce a generative model of data that mimics the structure of images while being analytically tractable. Natural images often display a hierarchical and compositional structure \cite{grenander1996elements}. Take, for example, the image of a snow leopard (see \autoref{fig:rhm}). This image is composed of multiple high-level components, such as the head and the paws. Each of these components, in turn, is composed of sub-features. For instance, the head comprises elements like ears, eyes, and mouth. Further dissecting these elements, we find even more granular details, such as edges that define the finer aspects of each feature. To model this hierarchical and compositional nature of images, we consider hierarchical generative models \cite{mossel2016deep,shalev2017failures, malach2018provably, malach2020implications,degiuli2019random,allen-zhu2023how,cagnetta2023deep}. 
In particular, consider a set of class labels $\mathcal{C}\equiv\left\lbrace 1,\dots, v\right\rbrace$ and an alphabet $\mathcal{A}\,{\equiv}\,\left\lbrace a_1,\dots,a_{v}\right\rbrace$ of $v$ features. Once the class label $\gamma$ is picked uniformly at random from $\mathcal{C}$, the data is generated iteratively from a set of production rules with branching factor $s$ at each layer $\ell$ (see ~\autoref{fig:rhm}, for an illustration):

\begin{eqnarray*}
\gamma \mapsto \mu^{(L-1)}_1,\dots,\mu^{(L-1)}_{s} \text{ for }\gamma\in\mathcal{C}\text{ and }\mu^{(L-1)}_i \in \mathcal{A},\\
\mu^{(\ell)} \mapsto \mu^{(\ell-1)}_{1},\dots,\mu^{(\ell-1)}_{s}\text{ for }\mu^{(\ell)}\in\mathcal{A},\mu^{(\ell-1)}_{i} \in \mathcal{A}, \\
\ell\in\{L-1,\dots,1\}.
\end{eqnarray*}

Since the total size of the data increases by a factor $s$ at each level, the input data are made of $d\equiv s^L$ input features $\mu^{(0)}$.
We adopt a one-hot encoding of these features, ultimately leading to a data vector $X\in \mathbb{R}^{d v}$. Note that for $\ell \geq 1$, the node variables correspond to latent variables, and there is no need to specify any choice of encoding.

For each level $\ell$, we consider that there are $m$ distinct production rules originating from the same higher-level feature $\mu^{(\ell)}$, i.e., there are $m$ equivalent lower-level representations of $\mu^{(\ell)}$. In addition, we assume that two distinct classes or latent variables cannot lead to the same low-level representation. This condition ensures, for example, that two distinct classes never lead to the same data.

We consider the case of the Random Hierarchy Model (RHM) \cite{cagnetta2023deep}, for which the $m$ production rules of any latent variable or class are sampled uniformly at random among the $v^s$ possible ones without replacement. In this case, the total number of possible data produced per class is $m\cdot m^s \cdots m^{s^{L-1}} = m^{\frac{d-1}{s-1}}$, which has exponential dependence in the dimension $d=s^L$.
In the following, we use the notation $X^{(\ell)}_i$ to indicate the variable at layer $\ell$ and position $i\in\{1, \dots, s^{L-\ell}\}$. 

In the context of unsupervised learning, a key parameter for this model is $f=m/v^{s-1}$. When $f=1$, all strings of latent variables of size $s$ can be produced at any level of the hierarchy. This implies that all possible $v^d$ input strings are generated, and the data distribution has little structure. When $f<1$, however, only a small fraction $\sim f^{(d-1)/(s-1)}$  of all possible strings is generated by the production rules. This implies that spatial correlations between different input positions appear, reflecting the hierarchy generating the data.

\hypertarget{sec:bp}{\section{Optimal denoising of the RHM with message passing}}
\label{sec:bp}

\begin{figure}
    \centering
    \begin{tikzpicture}
    \node[anchor=north west,inner sep=0pt] at (0,0){\resizebox{0.2\textwidth}{!}{\begin{tikzpicture}[varnode/.style={circle, draw=black, thick, minimum size=7mm, inner sep=0pt},
  factnode/.style={rectangle, draw=black, thick, minimum size=7mm, inner sep=0pt},
  level/.style={sibling distance=100mm/#1}, 
  edge from parent/.style={draw,thick}
  ]

\tikzset{
    varnode/.style={circle, draw, thick, minimum size=7mm, inner sep=0pt},
    factnode/.style={rectangle, draw, thick, minimum size=7mm, inner sep=0pt},
    midarrow/.style={decoration={
            markings,
            mark=at position 0.67 with {\arrow{Stealth}}
        },
        postaction={decorate},
        thick
    }
}
  
\node [varnode] (root) {$X_1^{(2)}$};
\node [factnode] (factor) [below= of root] {$\psi^{(2)}$};
\draw[midarrow] (factor) -- (root) node[midway, right] {$\nu_{\uparrow}(X_1^{(2)})$};

\node [varnode] (left) [below left=0.8cm and 0.5cm of factor] {$X_1^{(1)}$};
\draw[midarrow] (left) -- (factor) node[midway, left=0.1cm] {$\nu_{\uparrow}(X_1^{(1)})$};
\node [varnode] (center) [below right=0.8cm and 0.5cm of factor] {$X_2^{(1)}$};
\draw[midarrow] (center) -- (factor) node[midway, right=0.1cm] {$\nu_{\uparrow}(X_2^{(1)})$};

\node [factnode] (leftfactor1) [below =0.8cm of left] {$\psi^{(1)}$};
\draw[midarrow] (leftfactor1) -- (left);

\node [varnode] (leftvar1) [below left=0.8cm and 0cm of leftfactor1] {$X_{1}^{(0)}$};
\draw[midarrow] (leftvar1) -- (leftfactor1) node[midway, left=0.1cm] {$\nu_{\uparrow}(X_1^{(0)})$};
\node [varnode] (leftvar2) [below right=0.8cm and 0cm of leftfactor1] {$X_{2}^{(0)}$};
\draw[midarrow] (leftvar2) -- (leftfactor1);

\node [factnode] (rightfactor2) [below =0.8cm  of center] {$\psi^{(1)}$};
\draw[midarrow] (rightfactor2) -- (center);

\node [varnode] (rightvar1) [below left=0.8cm and 0cm of rightfactor2] {$X_{3}^{(0)}$};
\draw[midarrow] (rightvar1) -- (rightfactor2);
\node [varnode] (rightvar2) [below right=0.8cm and 0cm of rightfactor2] {$X_{4}^{(0)}$};
\draw[midarrow] (rightvar2) -- (rightfactor2) node[midway, right=0.1cm] {$\nu_{\uparrow}(X_4^{(0)})$};

\end{tikzpicture}}};
    \node[font=\sffamily\small] at (2ex,-2.5ex) {(Up)};
    \end{tikzpicture}
    \hspace{.1cm}
    \begin{tikzpicture}
    \node[anchor=north west,inner sep=0pt] at (0,0){\resizebox{0.2\textwidth}{!}{\begin{tikzpicture}[varnode/.style={circle, draw=black, thick, minimum size=7mm, inner sep=0pt},
  factnode/.style={rectangle, draw=black, thick, minimum size=7mm, inner sep=0pt},
  level/.style={sibling distance=100mm/#1}, 
  edge from parent/.style={draw,thick}
  ]

\tikzset{
    varnode/.style={circle, draw, thick, minimum size=7mm, inner sep=0pt},
    factnode/.style={rectangle, draw, thick, minimum size=7mm, inner sep=0pt},
    midarrow/.style={decoration={
            markings,
            mark=at position 0.67 with {\arrow{Stealth}}
        },
        postaction={decorate},
        thick
    }
}
  
\node [varnode] (root) {$X_1^{(2)}$};
\node [factnode] (factor) [below= of root] {$\psi^{(2)}$};
\draw[midarrow] (root) -- (factor) node[midway, right] {$\nu_{\downarrow}(X_1^{(2)})$};

\node [varnode] (left) [below left=0.8cm and 0.5cm of factor] {$X_1^{(1)}$};
\draw[midarrow] (factor) -- (left) node[midway, left=0.1cm] {$\nu_{\downarrow}(X_1^{(1)})$};
\node [varnode] (center) [below right=0.8cm and 0.5cm of factor] {$X_2^{(1)}$};
\draw[midarrow] (center) -- (factor) node[midway, right=0.1cm] {$\nu_{\uparrow}(X_2^{(1)})$};

\node [factnode] (leftfactor1) [below =0.8cm of left] {$\psi^{(1)}$};
\draw[midarrow] (left) -- (leftfactor1);

\node [varnode] (leftvar1) [below left=0.8cm and 0cm of leftfactor1] {$X_{1}^{(0)}$};
\draw[midarrow] (leftfactor1) -- (leftvar1) node[midway, left=0.1cm] {$\nu_{\downarrow}(X_1^{(0)})$};
\node [varnode] (leftvar2) [below right=0.8cm and 0cm of leftfactor1] {$X_{2}^{(0)}$};
\draw[midarrow] (leftvar2) -- (leftfactor1);

\node [factnode] (rightfactor2) [below =0.8cm  of center] {$\psi^{(1)}$};
\draw[midarrow] (rightfactor2) -- (center);

\node [varnode] (rightvar1) [below left=0.8cm and 0cm of rightfactor2] {$X_{3}^{(0)}$};
\draw[midarrow] (rightvar1) -- (rightfactor2);
\node [varnode] (rightvar2) [below right=0.8cm and 0cm of rightfactor2] {$X_{4}^{(0)}$};
\draw[midarrow] (rightvar2) -- (rightfactor2) node[midway, right=0.1cm] {$\nu_{\uparrow}(X_4^{(0)})$};

\end{tikzpicture}}};
    \node[font=\sffamily\small] at (3ex,-2.5ex) {(Down)};
    \end{tikzpicture}
    \caption{\textbf{Illustration of the flow of messages in the Belief Propagation algorithm for the case $s=2$, $L=2$ of the Random Hierarchy Model.} The factor nodes (squares) represent the rules that connect the variables at different levels of the hierarchy. The downward process is represented only for the leftmost branch.}
    \label{fig:BP}
\end{figure}
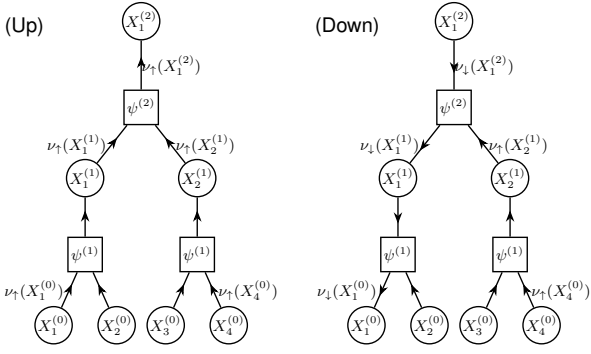

In this section, we characterize the Bayes optimal denoising process for the RHM. Given a noisy observation $X^{(0)}=x(t)$ of the input variables at time $t$, we compute $p(x(0)\vert x(t))$ exactly, obtaining full control of the statistics of the backward diffusion process from time $t$ to time $0$. In particular, given the tree structure of the model, we can compute the marginal probability of the values of all latent variables conditioned on $x(t)$ by using a message-passing algorithm. Therefore, we obtain the probability that a latent variable at level $\ell$ has changed when performing the forward-backward diffusion process for a duration $t$, a central quantity to interpret \autoref{fig:imagenet-main}.
The optimal denoising corresponds to reconstructing the data distribution $p(x(0))$ exactly. This perfect reconstruction corresponds to a diffusion model achieving perfect generalization. Although this is a strong assumption for modeling a neural network trained with empirical risk minimization, like the one considered in \Cref{sec:diffusion}, our theoretical analysis captures the phenomenology of our experiments.

\paragraph{Belief Propagation}
For computing the marginal distributions, we use Belief Propagation (BP) \cite{mossel2001reconstruction,  mezardmontanari}, which gives exact results for a tree graph such as the Random Hierarchy Model. In this case, the leaves of the tree correspond to the input variables at the bottom layer, and the root corresponds to the class variable at the top of the hierarchy. Each rule connecting variables at different levels corresponds to a factor node, as shown in \autoref{fig:BP}. 

The forward process adds noise to the variables in the input nodes. Each of these nodes sends its \textit{belief} on its value at $t=0$ to its parent latent node. These beliefs, or \textit{messages}, represent probabilistic estimates of the state of the sender node. Each latent node receives messages from all its children, updates its belief about its state, and sends its \textit{upward message} to its parent node. This process is repeated iteratively until the root of the tree. Subsequently, starting from the root, each node sends a \textit{downward message} to its children. Finally, the product of the upward and downward beliefs received at a given node represents the marginal probabilities of its state conditioned on the noisy observation. Hence, we can use these conditional marginals to compute the mean values of the variables at all levels of the hierarchy.
We assume that the production rules of the model are known by the inference algorithm, which corresponds to the optimal denoising process. 

The input variables $X^{(0)}$, in their one-hot-encoding representation, undergo the forward diffusion process of \autoref{eq:diffused-xt}, which can be defined in continuous time and constant $\beta_t$ by redefining $\overline{\alpha_t} = e^{-2t}$ and taking the limit $T\rightarrow\infty$ \cite{song2020score}. 

The denoising is made in two steps: the initialization of the messages at the leaves and the BP iteration.

\paragraph{initialization of the upward messages} 
In its one-hot-encoding representation, $X^{(0)}_i$ is a $v$-dimensional vector: taking the symbol $a_{\gamma}\in\{a_{1},\dots, a_{v}\}=\mathcal{A}$ corresponds to $X^{(0)}_{i} = e_{\gamma}$, with $e_{\gamma}$ a canonical basis vector.
Its continuous diffusion process takes place in $\mathbb{R}^{v}$: given the value $X^{(0)}_{i}=x_i(t)$, we can compute the probability of its starting value $p(x_i(0)\vert x_i(t))$ using Bayes formula. As derived in {\appBayes}, we obtain
    \begin{align}
        p({x_i(0) = e_{\gamma}} \vert x_i(t)) = \frac{1}{Z} \, e^{x_{i,\gamma}(t)/\Delta_t},
        \label{eq:bayes}
    \end{align}
with $\Delta_t = (1-\overline{\alpha_t})/\sqrt{\overline{\alpha_t}}$ and 
$Z = \sum_{\mu=1}^{v} e^{x_{i,\mu}(t)/\Delta_t}$.
This computation is performed independently for each input variable $i$, and therefore does not take into account the spatial correlations given by the generative model. The probabilities of \autoref{eq:bayes} are used to initialize the BP upward messages $\nnu{0} = p(x_i(0)\vert x_i(t))$ at the input variables.

\paragraph{BP iteration} Let $\rul{\ell}$ be any factor node connecting an $s$-tuple of low-level variables at layer $\ell-1$, $\{X_{i}^{(\ell-1)}\}_{i\in[s]}$, to a high-level variable $X_{1}^{(\ell)}$ at layer $\ell$. 
Without loss of generality, to lighten the notation, we rename the variables as $Y=X_{1}^{(\ell)}$, taking values $y\in\mathcal{A}$, and $X_i = X_{i}^{(\ell-1)}$, each taking values $x_i\in\mathcal{A}$.
For each possible association $y\rightarrow x_1,\dots, x_s$, the factor node $\rul{\ell}(y, x_1, ..., x_s)$ takes values
\beqs
    \rul{\ell}(y, x_1, ..., x_s) = \begin{cases}
        1, \; \text{if } y \rightarrow (x_1, ..., x_s) \text{ is rule at layer }\ell\\
        0, \; \text{otherwise}.
    \end{cases}
\eeqs
The BP upward and downward iterations for the (unnormalized) upward and downward messages respectively read
\beq
    \ncu{\ell+1}(y) = \sum_{x_1, ..., x_s \in \mathcal{A}^{\otimes s}} \rul{\ell+1}(y, x_1, ..., x_s) \prod_{i=1}^s \nnu{\ell}(x_i), \nonumber
\eeq
\begin{align}
    \ncd{\ell}(x_1) &= \sum_{\substack{x_2, ..., x_s \in \mathcal{A}^{\otimes (s-1)}\\ y\in \mathcal{A}}} \rul{\ell+1}(y, x_1, ..., x_s) \nonumber \\
    &\phantom{=} \times \nnd{\ell+1}(y)\prod_{i=2}^s \nnu{\ell}(x_i),
\end{align}
where $\nu_{\rho}^{(\ell)}(x) = \frac{\tilde{\nu}_{\rho}^{(\ell)}(x)}{\sum_{x'} \tilde{\nu}_{\rho}^{(\ell)}(x')}$, $\rho \in \{\uparrow,\downarrow\}$. The downward iteration, reported for $x_1$, can be trivially extended to the other variables $x_i$ by permuting the position indices. The values of $\nnu{0}(x_i)$ and $\nnd{L}(y)$ are set by the initial conditions. In particular, we initialize $\nnu{0}(x_i)$ as described in the previous paragraph and $\nnd{L}(y)=1/v$, which corresponds to a uniform prior over the possible classes $\mathcal{C}$.\footnote{This assumption corresponds to unconditioned diffusion, where the DDPM is not biased towards any specific class.}

\begin{figure}
    \centering
    \includegraphics[width=.9\columnwidth]{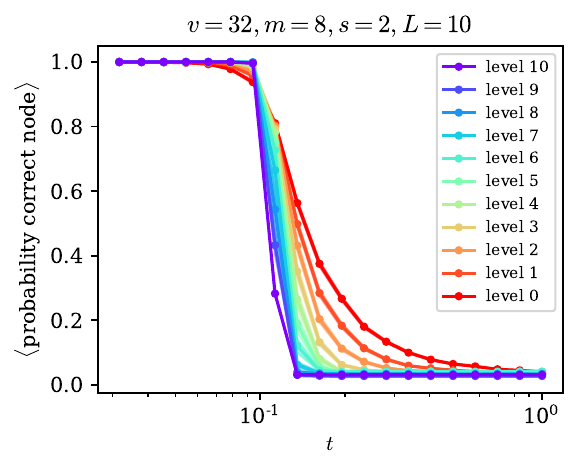}
    \vspace{-10pt}
    \caption{\textbf{Probability that the latent has not changed in the denoising process, corresponding to the largest marginal probability computed by BP, averaged for each layer, for varying inversion times of the diffusion process $t$.} Data for the RHM with $v=32$, $m=8$, $s=2$, $L=10$. Each level of the tree, indicated in the legend, is represented with a different color. We observe the same behavior of the curves for ImageNet data in \autoref{fig:imagenet-main}: the probability of the correct class has a sharp transition at a characteristic time scale, while the probabilities corresponding to latent variables in the lower levels change smoothly.}
    \label{fig:up-down-gauss}
\end{figure}

\paragraph{Results} We run the BP upward and backward iterations numerically. In \autoref{fig:up-down-gauss}, we show the probability corresponding to the correct symbol for each node of the tree. Remarkably, we note that \textit{(i)} the probability for the correct class at layer $L$ displays a transition at a characteristic time which becomes sharper for increasing $L$, and \textit{(ii)} the messages for the correct input variables and the correct latent variables at low levels of the tree change smoothly. In particular, the curves for messages at layer $L$ and layers $\ell < L$ invert their order at the transition, as in our observations on DDPMs and ImageNet data in \autoref{fig:imagenet-main}. This transition is one of our key findings, which we explain below.

\hypertarget{sec:mean-field}{\section{Mean-field theory of denoising diffusion}}
\label{sec:mean-field}

In this section, we make a simplifying assumption for the initial noise acting on the input and adopt a mean-field approximation to justify the existence of a phase transition. Remarkably, this approximation turns out to be of excellent quality for describing the diffusion dynamics. Specifically, consider a reference configuration at the leaves variables $X_i^{(0)} = \corr{x}_i$ that we would like to reconstruct, given a noisy observation of it. We assume that for each leaf variable, the noise is uniformly spread among the other symbols.\footnote{This is a mild approximation, as documented in {\appMapping}.} In other words, our belief in the correct sequence is corrupted by $\epsilon \in [0,1]$:

\begin{align}
\begin{cases}
    X_i^{(0)} = \corr{x}_i\quad \text{with belief } 1-\epsilon,\\
    X_i^{(0)} \text{ uniform over alphabet with belief } \epsilon.
\end{cases}
\end{align}

Hence, the initialization condition of the upward BP messages at a leaf node $X_i^{(0)}$ becomes 
\begin{align}
\begin{cases}
    \nnu{0}\lpa \corr{x}_i\rpa \qquad = 1-\epsilon+\epsilon/v,\\
    \nnu{0}\lpa x_i\neq\corr{x}_i \rpa = \epsilon/v,
\end{cases}
\label{eq:belief-main}
\end{align}
where $v$ is the alphabet cardinality.

Given these initial conditions and since the production rules are known, if $\epsilon=0$ -- i.e., in the noiseless case -- BP can reconstruct all the values of the latent variables exactly. Conversely, if $\epsilon=1$ -- i.e., when the input is completely corrupted and the belief on the leaves variables is uniform -- the reconstruction is impossible. In general, for a value of $\epsilon$, one is interested in computing the probability of recovering the latent structure of the tree at each layer $\ell$ and, as $L\to \infty$, to decide whether the probability of recovering the correct class of the input remains larger than $1/v$. 

\begin{figure}
    \centering
    \begin{tikzpicture}
    \node[anchor=north west,inner sep=0pt] at (0,0){
    \resizebox{.8\columnwidth}{!}{
    \input{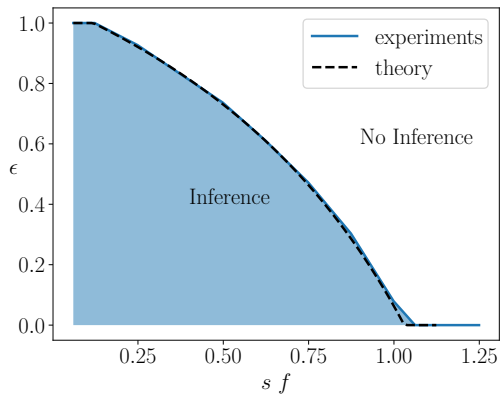}}
        };
    \end{tikzpicture}
    \caption{\textbf{Phase diagram for inferring the class node using the upward iteration of BP.} When $sf<1$, BP can infer the class if $\epsilon<\epsilon^*(sf)$. This transition is very well predicted by our theory. The inference region in the figure corresponds to the phase wherein the probability of the correct class is larger than the initialization belief in the correct values of the leaves, that is $1-\epsilon +\frac{\epsilon}{v}$. Experimental data are for a single realization of the RHM with $v=32$, $s=2$, $L=10$.}
    \label{fig:phase-diagram}
\end{figure}

\paragraph{Upward process} We begin by studying the upward process from the leaves. Consider a true input tuple $\corr{x}_1,\dots,\corr{x}_s$ which is associated with the higher-level feature $\corr{y}$. Given the randomness of the production rules, the messages are random variables depending on the specific realization of the rules. We adopt a \textit{mean-field} or \textit{annealed} approximation that neglects the fluctuations coming from the random choice of rules. Specifically, we approximate the upward message by the average upward message exiting the corresponding factor node $\langle \nnu{1}(y) \rangle_\rr$  over the possible realizations of $\rr$. In {\appBP}, we show that $\langle \nnu{1}(y) \rangle_\rr$ can take only two values: one for $y=\corr{y}$ and one for $y \neq \corr{y}$, as expected by symmetry considerations. Therefore, mean messages have the same structure as \autoref{eq:belief-main} and we can define a new $\epsilon'$. Introducing the probability of reconstructions $p = 1-\epsilon +\epsilon/v$ and $p' = 1-\epsilon' +\epsilon'/v$, we have

\beq
    p' = \frac{p^s + f \frac{m-1}{mv-1}\lpa 1-p^s\rpa}{p^s + f\lpa 1-p^s\rpa} = F(p).
    \label{eq:iterUP-main}
\eeq

\looseness=-1 Iterating this procedure across all the levels of the tree, we can compute the probability of recovering the correct class of the input. In particular, for large $L$, we are interested in studying the fixed points $p^*=F(p^*)$ of the iteration map in \autoref{eq:iterUP-main}. As derived in {\appUpAnneal}, when $sf>1$, this map has a repulsive fixed point $p^*=1$, which corresponds to $\epsilon = 0$, and an attractive fixed point $p^*=\nicefrac{1}{v}$, corresponding to $\epsilon = 1$. Thus, in this regime, inferring the class from the noisy observation of the input is impossible. In contrast, when $sf<1$, $p^*=1$ and $p^*=\nicefrac{1}{v}$ are both attractive fixed points, and a new repulsive fixed point $\nicefrac{1}{v} < p^* < 1$ separating the other two emerges. Therefore, in this second regime, there is a phase transition between a phase in which the class can be recovered and a phase in which it cannot. These theoretical predictions are numerically confirmed in the phase diagram in \autoref{fig:phase-diagram}.

Physically, $sf<1$ corresponds to a regime in which errors at lower levels of the tree do not propagate:  they can be corrected using information coming from neighboring nodes, thanks to the fact that only a small fraction of the strings are consistent with the production rules of the generative model. Conversely, when $sf>1$, even small corruptions propagate through the entire tree up to the root node and BP cannot infer the class correctly.

\paragraph{Downward process} The same calculation can be repeated for the downward process, with the additional difficulty that the downward iteration mixes upward and downward messages. We refer the reader to {\appBP} for the theoretical treatment.

\begin{figure*}
    \centering
    \begin{tikzpicture}
    \node[anchor=north west,inner sep=0pt] at (0,0){\resizebox{0.45\textwidth}{!}{\includegraphics[width=\columnwidth]{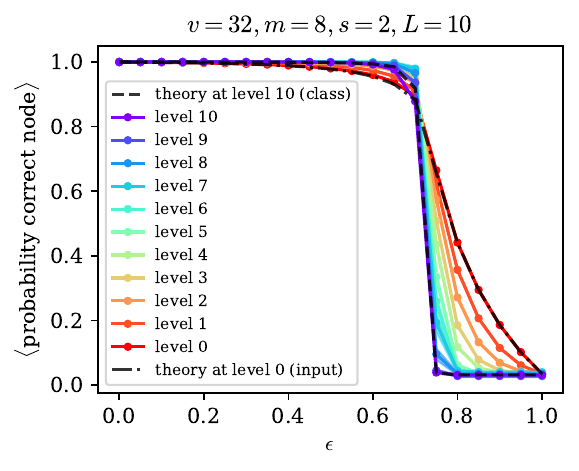}}};
    \node[font=\sffamily\large] at (2ex,-2.5ex) {(a)};
    \end{tikzpicture}
    \hspace{.1cm}
    \begin{tikzpicture}
    \node[anchor=north west,inner sep=0pt] at (0,0){\resizebox{0.45\textwidth}{!}{\includegraphics[width=0.98\columnwidth]{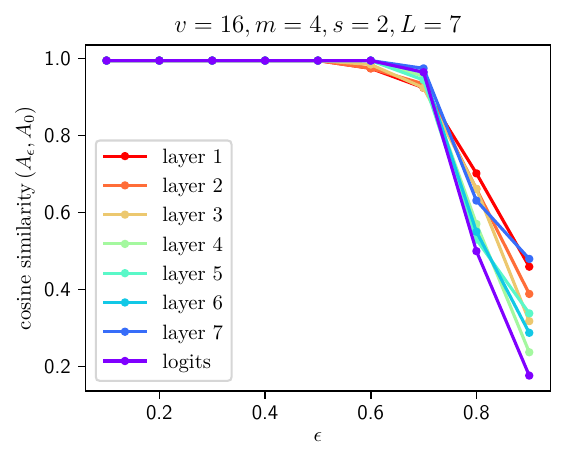}}};
    \node[font=\sffamily\large] at (3ex,-2.5ex) {(b)};
    \end{tikzpicture}
    \caption{\textbf{(a) Probability that the latent has not changed in the denoising process, corresponding to the largest marginal probability computed by BP, for varying $\epsilon$.} Data for the RHM with $v=32$, $m=8$, $s=2$, $L=10$. Each level of the tree, indicated in the legend, is represented with a different color. The black dashed lines are our mean-field theoretical predictions, which show excellent agreement with the experiments. In particular, the inversion between the curves for the top and bottom levels at the phase transition can be observed.
    \textbf{(b) Cosine similarity between the post-activations following every layer of a deep CNN trained on the RHM ($v=16$, $m=4$, $s=2$, $L=7$) for the starting and sampled data.} Each layer of the architecture, indicated in the legend, is represented with a different color. The curves showcase the same inversion predicted by our theory (cf. panel (a)).}
    \label{fig:activations_RHM}
\end{figure*}

\paragraph{Probabilities of reconstruction} Combining the mean upward and downward messages, we obtain a theoretical prediction for the probabilities of reconstructing the correct values of the variables at each layer. We compare our theoretical predictions with numerical experiments in \autoref{fig:activations_RHM}-(a). In these experiments, BP equations are solved exactly for a given RHM starting with the initialization of \autoref{eq:belief-main}. Our theory perfectly captures the probability of reconstruction for the input nodes and the class. Moreover, in {\appBP} we show that our theory predicts the probabilities of reconstruction of latent nodes at all layers. 

\paragraph{Experiment on CNN's activations} Similarly to our experiment on the ConvNeXt in \Cref{sec:diffusion}, we investigate how the hidden representation of a model trained to classify the RHM changes when its input is denoised starting from a corruption noise $\epsilon$. We consider an instantiation of the RHM with $L=7$, $s=2$, $v=16$, and $m=4$. First, we train a convolutional neural network with $L=7$ layers, matching the tree structure of the model, with $n=300k$ training examples up to interpolation. The resulting architecture has 99.2\% test accuracy. To sample new data from noisy observations of held-out data, we start by sampling the root using the marginal probability computed with BP. Then, we update the beliefs and the marginals conditioning on the sampled class, and sample one latent variable at layer $L-1$. We iterate this procedure node-by-node, descending the tree until we obtain a sampled configuration at the bottom layer \cite{mezardmontanari}. For each corrupting noise $\epsilon$ and each layer of the CNN, we compute the cosine similarity between post-activations for the initial and generated configurations. Panel (b) of \autoref{fig:activations_RHM} shows the obtained curves. Remarkably, we observe the same qualitative behavior as in panel (a) of \autoref{fig:activations_RHM}, ultimately explaining the empirical observation of \autoref{fig:imagenet-main}.


\section{Conclusions}

We have argued that reversing time in denoising diffusion models opens a window on the compositional nature of data. For synthetic hierarchical generative models of data, where the Bayes optimal denoising can be exactly computed, low-level features can already change at small times, but the class remains most often the same. At larger times, a phase transition is found where the probability of remaining in the same class suddenly drops to random chance.  Yet,  low-level features identical to those of the initial sample can persist and compose the new sample. 
Strikingly, this theoretical analysis characterizes well the results found with ImageNet, where the denoising is performed by a trained U-Net. Interestingly, the structure of the U-Net with the skip connections between the encoder and decoder parts mimics the upward and downward iterations of belief propagation, where the downward process mixes upward and downward messages. In fact, building on the present work \cite{mei2024unets} shows that U-Nets are capable of effectively approximating the belief propagation denoising algorithm. Investigating whether the function learned by U-Nets approximates BP is a promising avenue for future work.
In the present work, we used the internal representation of deep networks as a proxy for the hierarchical structure of images. An interesting direction for future work will be using deep hierarchical segmentation techniques \cite{arbelaez2010contour,Ge_2023_CVPR, xie2021unsupervised, zhang2020self}
to extract latent variables, so as to test our predictions on their evolution in forward-backward experiments.
Finally, future work can test our theoretical predictions on other modalities successfully handled by diffusion models, such as language and biological structures.

The interplay between the hierarchy in feature space and in time revealed here may help understand the puzzling success of diffusion models, including the number of data needed to train such methods, or why they can generalize and not simply memorize the empirical distribution on which they were trained \cite{somepalli2022diffusion,carlini2023extracting,yoon2023diffusion}. More generally, our results put forward hierarchical generative models as tools to understand open questions for other methods, ranging from the emergence of new skills by the composition of more elementary ones in foundation models to that of transferable representations in self-supervised learning.

\matmethods{
The code for reproducing the experiments is available at \href{https://github.com/pcsl-epfl/forward-backward-diffusion}{github.com/pcsl-epfl/forward-backward-diffusion} \cite{sclocchiGitHub}.
}

\acknow{We thank Francesco Cagnetta and Umberto Maria Tomasini for helpful discussions. We thank Guillermo Ortiz-Jiménez and Stefano Sarao for feedback on the manuscript. This work was supported by a grant from the Simons Foundation (\#454953 Matthieu Wyart).
}
\showmatmethods{} 
\showacknow{} 

\bibsplit[15]

\bibliography{bibliography}

\newpage
\onecolumn

\appendixstyle

\SItext

\hypertarget{app:bayes}{\section{Belief Propagation initialization for the denoising of the RHM}}
\label{app:bayes}

As discussed in \secBP, we define the diffusion process for the input variable $X_{i}^{(0)}$ in the space $\mathbb{R}^v$. In particular, its value $x(t)$ at time $t$ is

\begin{align}
    x(t) = \sqrt{\overline{\alpha_t}} x(0) + \sqrt{1-\overline{\alpha_t}} \eta,
    \label{eq:app_diffusion}
\end{align}
with $\eta \sim \mathcal{N}(0,\mathbb{I}_{v})$ and $x(0)$ its starting value at time $t$, which is a one-hot-encoding vector of the form $x(0) = e_{\mu}$. Given the value $x(t)$, the conditional probabilities for the values of $x(0)$ are given by Bayes rule
\begin{align}
    p\lpa x(0) = e_{\mu} \vert x(t)\rpa = 
    \frac{p\lpa x(t)\vert x(0) = e_{\mu}\rpa \, p\lpa x(0) = e_{\mu}\rpa}{\sum_{\lambda} p\lpa x(t)\vert x(0) = e_{\lambda}\rpa \, p\lpa x(0) = e_{\lambda}\rpa}.
\label{eq:app_bayes1}    
\end{align}

The prior probabilities on $x(0)$ are taken to be uniform over the alphabet, i.e., $p\lpa x(0) = e_{\lambda}\rpa = 1/v$, $\forall \lambda$, while $p\lpa x(t)\vert x(0) = e_{\mu}\rpa$ is given by the diffusion process of \autoref{eq:app_diffusion}:
\begin{align}
\begin{split}
    p\lpa x(t)\vert x(0)= e_{\mu}\rpa 
    &= C_{t} \exp\lsq-\frac{1}{2(1-\overline{\alpha_t})}\sum_{\gamma}\lpa x_\gamma(t) - \sqrt{\overline{\alpha_t}} e_{\mu} \rpa^2 \rsq =\\
    &=C_t \exp\lsq- \frac{\|x(t)\|^2 + \overline{\alpha_t}}{2(1-\overline{\alpha_t})} \rsq
    \exp\lsq \frac{\sqrt{\overline{\alpha_t}} }{1-\overline{\alpha_t}} x_\mu(t) \rsq, 
\label{eq:app_bayes2}
\end{split}
\end{align}
where $C_t$ is the normalization constant. Putting \autoref{eq:app_bayes2} into \autoref{eq:app_bayes1}, we obtain
\begin{align}
    p\lpa x(0) = e_{\mu} \vert x(t)\rpa = 
    \frac{1}{Z} e^{\frac{\sqrt{\overline{\alpha_t}} }{1-\overline{\alpha_t}} x_\mu(t)},
\end{align}
with $Z= \sum_{\lambda=1}^v e^{\frac{\sqrt{\overline{\alpha_t}} }{1-\overline{\alpha_t}} x_\lambda(t)}$.

\hypertarget{app:bp}{\section{Belief Propagation equations}}
\label{app:bp}

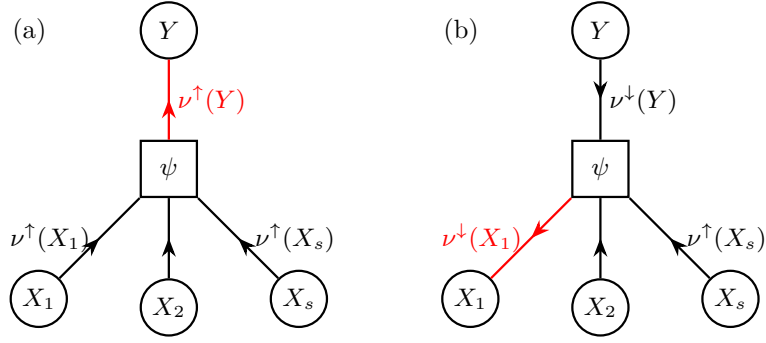
\begin{figure}[H]
    \centering
    \begin{tikzpicture}
        \node[anchor=north west,inner sep=0pt] at (0,0){
        \resizebox{.25\textwidth}{!}{\begin{tikzpicture}[varnode/.style={circle, draw=black, thick, minimum size=7mm, inner sep=0pt},
  factnode/.style={rectangle, draw=black, thick, minimum size=7mm, inner sep=0pt},
  level/.style={sibling distance=100mm/#1}, 
  edge from parent/.style={draw,thick}
  ]

\tikzset{
    varnode/.style={circle, draw, thick, minimum size=7mm, inner sep=0pt},
    factnode/.style={rectangle, draw, thick, minimum size=7mm, inner sep=0pt},
    midarrow/.style={decoration={
            markings,
            mark=at position 0.5 with {\arrow{Stealth}}
        },
        postaction={decorate},
        thick
    }
}
  
\node [varnode] (root) {$Y$};
\node [factnode] (factor) [below= of root] {$\psi$};
\draw[midarrow, red] (factor) -- (root) node[midway, right=0.1cm] {$\nu^{\uparrow}(Y)$};

\node [varnode] (left) [below left= of factor] {$X_1$};
\draw[midarrow] (left) -- (factor) node[midway, left=0.1cm] {$\nu^{\uparrow}(X_1)$};
\node [varnode] (center) [below = of factor] {$X_2$};
\draw[midarrow] (center) -- (factor);
\node [varnode] (right) [below right= of factor] {$X_s$};
\draw[midarrow] (right) -- (factor) node[midway, right=0.2cm] {$\nu^{\uparrow}(X_s)$};

\end{tikzpicture}}};
        \node[font=\large] at (2ex,-3ex) {(a)};
    \end{tikzpicture}
    \hspace{1cm}
    \begin{tikzpicture}
        \node[anchor=north west,inner sep=0pt] at (0,0){
        \resizebox{.25\textwidth}{!}{\begin{tikzpicture}[varnode/.style={circle, draw=black, thick, minimum size=7mm, inner sep=0pt},
  factnode/.style={rectangle, draw=black, thick, minimum size=7mm, inner sep=0pt},
  level/.style={sibling distance=100mm/#1}, 
  edge from parent/.style={draw,thick}
  ]

\tikzset{
    varnode/.style={circle, draw, thick, minimum size=7mm, inner sep=0pt},
    factnode/.style={rectangle, draw, thick, minimum size=7mm, inner sep=0pt},
    midarrow/.style={decoration={
            markings,
            mark=at position 0.5 with {\arrow{Stealth}}
        },
        postaction={decorate},
        thick
    }
}
  
\node [varnode] (root) {$Y$};
\node [factnode] (factor) [below= of root] {$\psi$};
\draw[midarrow] (root) -- (factor) node[midway, right=0.1] {$\nu^{\downarrow}(Y)$};

\node [varnode] (left) [below left= of factor] {$X_1$};
\draw[midarrow, red] (factor) -- (left) node[midway, left=0.1cm] {$\nu^{\downarrow}(X_1)$};
\node [varnode] (center) [below = of factor] {$X_2$};
\draw[midarrow] (center) -- (factor);
\node [varnode] (right) [below right= of factor] {$X_s$};
\draw[midarrow] (right) -- (factor) node[midway, right=0.2cm] {$\nu^{\uparrow}(X_s)$};

\end{tikzpicture}}};
        \node[font=\large] at (2ex,-3ex) {(b)};
    \end{tikzpicture}
    \caption{Factor tree-graph connecting the higher-level feature $Y$ to the lower-level features $X_{i=1,...,s}$ according to the rules $\rr$. 
    The upward messages $\nu^{\uparrow}(y)$ are computed from the upward messages $\nu^{\uparrow}(x_i)$ coming from the 
    nodes $X_i$, connected to $Y$ through the rule $\rr$ (panel (a)).
    The downward messages $\nu^{\downarrow}(x_1)$, instead, are computed from both the downward messages $\nu^{\downarrow}(y)$ coming from $Y$ and the upward messages $\nu^{\uparrow}(x_i)$ coming from the nodes $X_{i=2,...s}$, connected to $X_1$ through the rule $\rr$ (panel (b)).
    }
    \label{fig:patch}
\end{figure}

Given a factor tree-graph, the Belief Propagation (BP) equations compute iteratively the messages going from the variable nodes to the factor nodes and vice-versa, starting from the initialization conditions at the leaves and root of the tree-graph \cite{mezardmontanari}. For the generative model defined in \secRHM, the leaves correspond to the variables at the bottom layer while the root is the class variable at the top of the hierarchy. 
Each rule, connecting variables at different layers, corresponds to a factor node. The BP messages that flow from the variable nodes to the factor nodes, therefore, correspond to upward messages, while those going from factor nodes to variables correspond to downward messages (\autoref{fig:patch}).

To each variable node $X^{(\ell)}_i$ at level $\ell$, we associate the upward messages $\nnu{\ell}$ and downward messages $\nnd{l}$, one for each possible value of the alphabet it can take.
To simplify the notation, here we consider how messages propagate from one level to the other, and we call $Y$ the variable corresponding to the higher level and $X_{i=1,...,s}$ the lower level ones connected to it. The factor node connecting them is such that, for each possible association $y\rightarrow x_1,\dots, x_s$, it takes values
\beqs
    \rul{\ell}(y, x_1, ..., x_s) = \begin{cases}
        1, \quad \text{if } y \rightarrow (x_1, ..., x_s) \text{ is a rule at layer }\ell\\
        0, \quad \text{otherwise}.
    \end{cases}
\eeqs

The BP upward and downward iterations are defined as follows.
\begin{itemize}
    \item Upward iteration:
\beq
    \ncu{\ell+1}(y) = \sum_{x_1, ..., x_s \in \mathcal{A}^{\otimes s}} \rul{\ell+1}(y, x_1, ..., x_s) \prod_{i=1}^s \nnu{\ell}(x_i)\ ,
    \label{eq:nuUP}
\eeq
\beq
    \nnu{\ell}(y) = \frac{\ncu{\ell}(y)}{\sum_{y'} \ncu{\ell}(y')}.
\eeq

    \item Downward iteration:
\beq
    \ncd{\ell}(x_1) = \sum_{\substack{x_2, ..., x_s \in \mathcal{A}^{\otimes (s-1)}\\ y\in \mathcal{A}}} \rul{\ell+1}(y, x_1, ..., x_s)\ \nnd{\ell+1}(y)\prod_{i=2}^s \nnu{\ell}(x_i)\
    \label{eq:nuDOWN}
\eeq
\beq
    \nnd{l}(x) = \frac{\ncd{\ell}(x)}{\sum_{x'} \ncd{\ell}(x')}.
\eeq
\end{itemize}

$\nnu{\ell}(y)$ and $\nnd{\ell}(x)$ are fluctuating quantities that depend on the position of the node.

\hypertarget{app:anneal}{\subsection{$\epsilon$-process}}

In this process, we consider a reference configuration at the leaves variables $X_i^{(0)} = \corr{x}_i$ that we would like to reconstruct, given a noisy observation of it. As a result of this noise addition, our belief in the correct sequence is corrupted by $\epsilon \in [0,1]$:
\begin{align}
\begin{cases}
    X_i^{(0)} = \corr{x}_i\quad \text{with belief } 1-\epsilon\\
    X_i^{(0)} \text{ uniform over alphabet with belief } \epsilon.\\
\end{cases}
\end{align}

Therefore, the initialization condition of the upward BP messages at a leaf node $X_i^{(0)}$ is 
\begin{align}
\begin{cases}
    \nu_{\uparrow}\lpa \corr{x}_i\rpa \qquad = 1-\epsilon+\epsilon/v,\\
    \nu_{\uparrow}\lpa x_i\neq\corr{x}_i \rpa = \epsilon/v, \\
\end{cases}
\label{eq:belief}
\end{align}
where $v$ is the corresponding alphabet size.\\
The initialization condition at the root node $X^{(L)}$, that corresponds to the messages $\nnd{L}$ for that node, is uniform over the alphabet $\mathcal{A}$, so that the algorithm has no bias on any specific class.

\subsubsection{Upward iteration}
\label{app:upward_anneal}

We consider the upward iteration when going from the bottom layer to the one above it. Let $X_1, \dots, X_s$ denote a tuple at the bottom level which is associated with the reference values $\corr{x}_i$. This tuple is connected to the higher level variable $Y$ via a set of rules $\rr$ (\autoref{fig:patch}). According to $\rr$, the association from the high-level feature to the reference low-level sequence $\corr{x}_1,\dots,\corr{x}_s$ is given by
\beqs
     \corr{y}\rightarrow \corr{x}_1,\dots,\corr{x}_s.
\eeqs

We call $\ham{\mathbf{w}, \mathbf{z}}$ the Hamming distance between two sequences $\mathbf{w} = [w_1, \dots, w_s]$, $\mathbf{z}=[z_1, \dots, z_s]$ of length $s$.\\
From \autoref{eq:belief}, at the bottom layer, the belief in a sequence $\x = [x_1,\dots, x_s]$ with $\ham{\x,\corr{\x}} = k \in\{0,...,s\}$ from $\corr{\x} = [\corr{x}_1,\dots, \corr{x}_s]$ is
\beq
    \bel(k) = \lpa\frac{\epsilon}{v}\rpa^{k} \lpa 1-\epsilon+\frac{\epsilon}{v}\rpa^{s-k}
    \label{eq:bel_k}
\eeq

The non-normalized upward messages for the variable $Y$ are given by:
\beq
    \tilde{\nu}_{\uparrow}(y) = 
    \sum_{x_1, ..., x_s} \rr(y, x_1, ..., x_s) \prod_{i=1}^s \nu_{\uparrow}(x_i) = 
    \sum_{\x\in\samp} \rr(y, \x) \bel\lpa\ham{\x,\corr{\x}}\rpa,
    \label{eq:nu_y}
\eeq
where we are using the short-hand notation $\rr(y, x_1, ..., x_s) = \rr(y, \x)$ and we have restricted the sum over the set $\samp$ of sequences $\x$ that appear in the possible rules $y\rightarrow x_1,\dots, x_s$. In fact, if $\x\notin\samp$, then $\rr(y, \x) = 0$.

For $\x\in\samp$, the factor $\rr(y, \x)$ is such that:
\begin{itemize}
    \item if $\ham{\x,\corr{\x}}=0$:
    \beq
    \begin{aligned}
        \begin{cases}
            &\rr(\corr{y}, \corr{x}_1, ..., \corr{x}_s) = 1,\\
            &\rr(y, \corr{x}_1, ..., \corr{x}_s) = 0, \quad y\neq \corr{y}.\\
        \end{cases}
    \end{aligned}
    \eeq
    \item if $\ham{\x,\corr{\x}}>0$:
    \beq
    \begin{aligned}
        \begin{cases}
            &\rr(\tilde{y}, x_1, ..., x_s) = 1, \quad \text{for some $\tilde{y}$ independent of $\corr{y}$}\\
            &\rr(y, x_1, ..., x_s) = 0, \quad y\neq \tilde{y}.\\
        \end{cases}
    \label{eq:randomPsi}
    \end{aligned}
    \eeq    
\end{itemize}

We can decompose \autoref{eq:nu_y} as
\beq
    \tilde{\nu}_{\uparrow}(y) = 
    \delta_{y,\corr{y}} \bel(0) +
    \sum_{k=1}^s
    \bel\lpa k\rpa
    \lsq
    \sum_{\substack{\x\in\samp\\
    \ham{\x,\corr{\x}}=k
    }} \psi(y, \x)
    \rsq.
\eeq

\paragraph{Annealed average}
$\psi$ is a random quantity and we want to compute the average message $\langle \tilde{\nu}_{\uparrow}(y)\rangle_\rr$ over the possible realizations of $\rr$. We can decompose the selection of the rules in two steps: sampling the set of $mv-1$ sequences $\{\x, \x \neq \corr{\x}\}$ and then associating the $v$ higher-level features $y$ to them. Therefore, for a generic quantity $A$, we indicate the average over the rules realization $\langle A \rangle_\rr$ as $\langle\langle A \rangle_{\asso}\rangle_{\samp}$, where $\langle\dots\rangle_{\samp}$ is the average over the sequence sampling and $\langle \dots\rangle_{\asso}$ is the average over the $y\leftarrow \x$ associations:

\beq
    \langle \tilde{\nu}_{\uparrow}(y) \rangle_\psi= 
    \delta_{y,\corr{y}} \bel(0) +
    \sum_{k=1}^s
    \bel\lpa k\rpa
    \langle\langle
    \sum_{\substack{\x\in\samp\\
    \ham{\x,\corr{\x}}=k
    }} \psi(y, \x)
    \rangle_{\asso}\rangle_{\samp}
\eeq

Since for each sequence $\x\neq \corr{\x}$ the association $y\leftarrow \x$ is done randomly, independently of $\ham{\x,\corr{\x}}$, then from \autoref{eq:randomPsi}, we have $\langle \psi(y, \x) \rangle_{\asso} \simeq 1/v$. More precisely, since we have associated the reference sequence $\corr{\x}$ to $\corr{y}$:
\beq
    \overline{\psi_y} = \langle \psi(y, \x) \rangle_{\asso} = 
    \frac{m-1}{mv-1} \delta_{y,\corr{y}} + \frac{m}{mv-1} \lpa 1-\delta_{y,\corr{y}} \rpa.
\eeq

Therefore:
\beq
\begin{aligned}
    \langle \tilde{\nu}_{\uparrow}(y) \rangle_\psi &= 
    \delta_{y,\corr{y}} \bel(0) +
    \overline{\psi_y}
    \sum_{k=1}^s
    \bel\lpa k\rpa
    \langle
    \sum_{\substack{\x\in\samp\\
    \ham{\x,\corr{\x}}=k
    }} 1
    \rangle_{\samp} = \\
    & = \delta_{y,\corr{y}} \bel(0) +
    \overline{\psi_y}
    \sum_{k=1}^s
    \bel\lpa k\rpa
    \langle
    n_k
    \rangle_{\samp},
\end{aligned}
\eeq

where $n_k$ is the number of sequences $\x\in\samp$ having Hamming distance $\ham{\x,\corr{\x}} = k$ from $\corr{\x}$. Since the sequences are sampled randomly, the numbers $n_1, ..., n_s$ are distributed according to a multivariate hyper-geometric distribution,
\beq
    P\lpa n_1, ..., n_s \rpa = \frac{\prod_{k=1}^s \binom{\binom{s}{k}(v-1)^k}{n_k}}{\binom{v^s-1}{mv-1}},
\eeq
which gives the averages 
\beq
    \langle n_k \rangle_{\samp} = \frac{mv-1}{v^s-1} \binom{s}{k}(v-1)^k = f \binom{s}{k}(v-1)^k,
\eeq
with
\beq
    f = \frac{mv-1}{v^s-1}.
\eeq

Therefore:
\beq
\begin{aligned}
    \langle \tilde{\nu}_{\uparrow}(y) \rangle_\psi = 
    \delta_{y,\corr{y}} \bel(0) +
    f\ \overline{\psi_y}
    \sum_{k=1}^s
    \bel\lpa k\rpa
    \binom{s}{k}(v-1)^k
    .
\end{aligned}
\eeq

From the beliefs \autoref{eq:bel_k}, we see that
\beq
    \sum_{k=1}^s
    \bel\lpa k\rpa
    \binom{s}{k}(v-1)^k=
    \sum_{k=1}^s
    \binom{s}{k}(v-1)^k
    \lpa\frac{\epsilon}{v}\rpa^{k} \lpa 1-\epsilon+\frac{\epsilon}{v}\rpa^{s-k} = 
    \lsq 1 - \lpa 1-\epsilon+\frac{\epsilon}{v}\rpa^s \rsq
    = 1-\bel(0),
\eeq

which gives
\beq
\begin{aligned}
    \langle \tilde{\nu}_{\uparrow}(y) \rangle_\psi = 
    \delta_{y,\corr{y}} \bel(0) +
    f\ \overline{\psi_y}
    \lsq 1- \bel(0) \rsq
\end{aligned}
.
\eeq

The normalization constant is:
\beq
    \langle Z_{\uparrow} \rangle_\rr = \sum_{y} \langle \tilde{\nu}_{\uparrow}(y) \rangle_\rr = \bel(0) + f \lsq 1- \bel(0) \rsq .
\eeq

Finally, we obtain the average belief for $Y$

\beq
    \langle \nu_{\uparrow}(y) \rangle_\psi = \frac{\langle \tilde{\nu}_{\uparrow}(y) \rangle_\rr}{\langle Z_{\uparrow} \rangle_\rr} = 
    \frac{\delta_{y,\corr{y}} \bel(0) + f\ \overline{\psi_y} \lsq 1- \bel(0) \rsq}{\bel(0) + f \lsq 1- \bel(0) \rsq}
\eeq
 We have that:
\begin{itemize}
    \item for $y=\corr{y}$
    \beq
        \langle \nu_{\uparrow}(\corr{y}) \rangle_\psi =
        \frac{\bel(0) + f \frac{m-1}{mv-1} \lsq 1- \bel(0) \rsq}{\bel(0) + f \lsq 1- \bel(0) \rsq},
        \label{eq:nu_corr}
    \eeq
    \item for $y\neq \corr{y}$
    \beq
        \langle \nu_{\uparrow}(y) \rangle_\psi =
        f \frac{m}{mv-1} \frac{1- \bel(0)}{\bel(0) + f \lsq 1- \bel(0) \rsq}.
        \label{eq:nu_mist}
    \eeq
\end{itemize}

\paragraph{Iterating over layers}
The average messages in \autoref{eq:nu_corr}, \autoref{eq:nu_mist} are of two kinds: one for the reference feature $\corr{y}$ and another for the others $y\neq \corr{y}$, and they both depend on the previous beliefs through $\bel(0) = \lpa 1-\epsilon + \frac{\epsilon}{v}\rpa^s$. Therefore, the average messages at the higher level have the same structure as those at the lower level \autoref{eq:belief}. We can then define a new $\epsilon'$:
\beq
    1-\epsilon' + \frac{\epsilon'}{v} = \frac{\lpa 1-\epsilon + \frac{\epsilon}{v}\rpa^s + f \frac{m-1}{mv -1}\lsq 1- \lpa 1-\epsilon + \frac{\epsilon}{v}\rpa^s \rsq }{\lpa 1-\epsilon + \frac{\epsilon}{v}\rpa^s + f \lsq 1 - \lpa 1-\epsilon + \frac{\epsilon}{v}\rpa^s \rsq}
\eeq
or, equivalently, 
\beq
    p' = \frac{p^s + f \frac{m-1}{mv-1}\lpa 1-p^s\rpa}{p^s + f\lpa 1-p^s\rpa} = F(p)
    \label{eq:iterUP}
\eeq
with $p' = 1-\epsilon' +\epsilon'/v$ and $p = 1-\epsilon +\epsilon/v$. The derivative of $F(p)$ with respect to $p$ is given by
\beq
    F'(p) = \frac{m(v-1)}{mv -1}\ \frac{f s p^{s-1}}{\lsq p^s + f(1-p^s) \rsq^2}=
    fs \frac{p^{s-1}}{\lsq p^s + f(1-p^s) \rsq^2} + \O\lpa\frac{1}{v}\rpa
\eeq

We can extend the tree in \autoref{fig:patch} iteratively to higher levels of the hierarchy, where the variables $Y$ take the place of the variables $X_i$ and so on. 

The iteration \autoref{eq:iterUP} has fixed points $p= 1$ (corresponding to $\epsilon=0$) or $p=1/v$ (corresponding to $\epsilon=1$).
An additional repulsive fixed point at finite $p$ appears if
\beq
    F'(1)<1,
\eeq
that is
\beq
    \frac{m(v-1)}{mv -1}\ f s < 1
    \quad \Rightarrow\quad fs < 1 + \frac{1-1/m}{v-1}   
\eeq
\beq
\boxed{
\Rightarrow\quad fs < 1 + \O\lpa\frac{1}{v}\rpa
}\,.
\eeq

\begin{figure}[H]
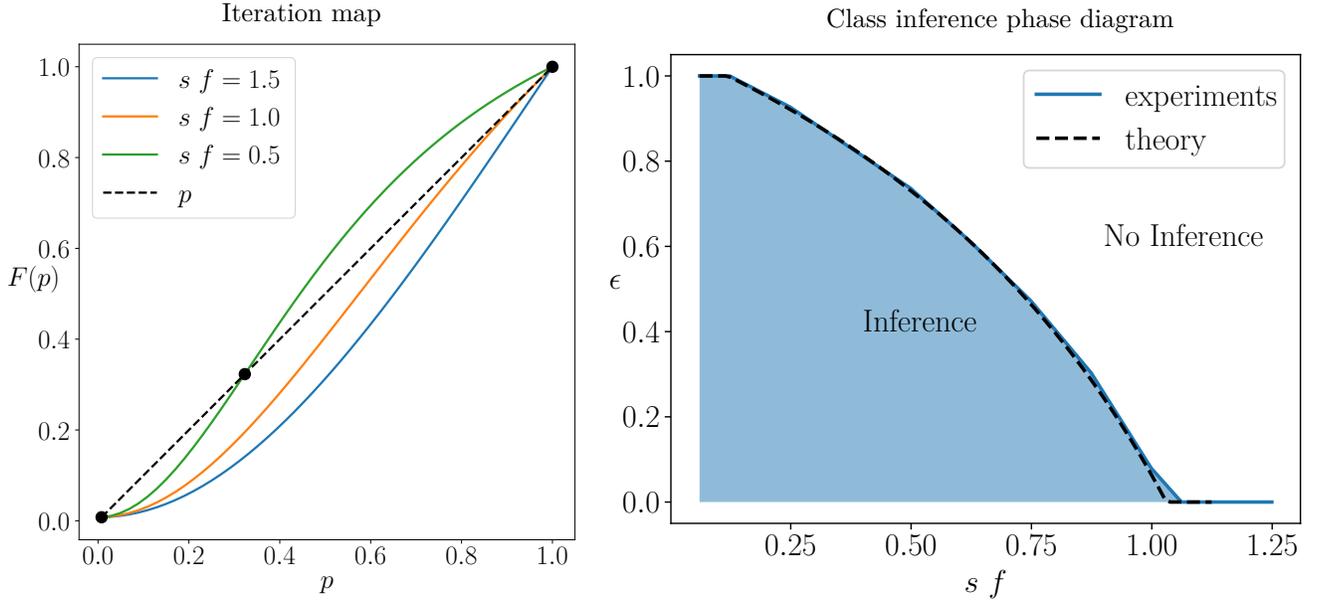

    \centering
    \begin{tikzpicture}
    \node[font=\large] at (30ex,2ex) {Iteration map};
    \node[anchor=north west, inner sep=0pt] at (0,0){
    \resizebox{.44\columnwidth}{!}{
    \input{figures/BP_iteration.pgf}}
        };
    \end{tikzpicture}
    \begin{tikzpicture}
    \node[font=\large] at (40ex,2ex) {Class inference phase diagram};
    \node[anchor=north west,inner sep=0pt] at (0,0){
    \resizebox{.54\columnwidth}{!}{
    \input{figures/BP_upward-root_inference-v=32_s=2_L=10.pgf}}
        };
    \end{tikzpicture}
    \caption{
    \textit{Left panel:} 
    \textbf{Iteration map of \autoref{eq:iterUP}.} For $s f>1$, there are only two fixed points $p=F(p)$ corresponding to $p_1=1$ (repulsive) and $p_2=1/v$ (attractive). For $s f<1$, there is another (repulsive) fixed point at finite $p^*$, $1/v<p^*<1$, separating $p_1$ and $p_2$ (both attractive).
    \textit{Right panel:} 
    \textbf{Phase diagram for inferring the class node using the upwards iteration of BP.} When $sf<1$, BP can infer the class of the datum if $\epsilon<\epsilon^*(sf)$. This transition is well predicted by $p^*=1-\epsilon^*+\epsilon^*/v$, with $p^*=F(p^*)$ from \autoref{eq:iterUP}. Experimental data for $v=32$, $s=2$, $L=10$.
    }
    \label{fig:enter-label}
\end{figure}

\subsubsection{Downward iteration}
\label{app:downward_anneal}

We consider the downward process when we try reconstructing the reference association $\corr{y}\rightarrow \corr{x}_1,\dots,\corr{x}_s$ from higher-level variable $Y$ to the corresponding lower-level tuple $X_1, \dots, X_s$, via the set of rules $\rr$.
We consider the downward message received by the variable $X_1$ (\autoref{fig:patch}):
\begin{align}
\begin{split}
    \tilde{\nu}_{\downarrow}(x_1) &= \sum_{\substack{x_2, ..., x_s \in \mathcal{A}^{\otimes (s-1)}\\ y\in \mathcal{A}}} \rr(y, x_1, ..., x_s)\ \nu_{\downarrow}(y) \prod_{i=2}^s \nu_{\uparrow}(x_i)=\\
    &=\delta_{x_1, \corr{x}_1} \nu_{\downarrow}(\corr{y})\prod_{i=2}^s \nu_{\uparrow}(\corr{x}_i)
    +
    \sum_{y}
    \nu_{\downarrow}(y)
    \sum_{\substack{x_2,\dots,x_s\\
    \x\in\samp, \ham{\x,\corr{\x}}>0
    }}
    \psi(y, \x)
    \prod_{i=2}^s \nu_{\uparrow}(x_i)
\end{split}
\label{eq:nu_down}
\end{align}

\paragraph{Annealed average}
To study the iteration of \autoref{eq:nu_down} analytically, we compute the average message $\langle \tilde{\nu}_{\downarrow}(x_1)\rangle_\rr$ over the realizations of the random rules $\rr$ as done in \Cref{app:downward_anneal} for the upward iteration. 

We call $n_{x_1}$ the number of sequences, having $X_1=x_1$, that have been sampled by the choice of the rules.
The numbers $n_{x_1}$ are distributed according to a multivariate hyper-geometric distribution,
\beq
    P\lpa \{n_{x_1}\}_{x_1\in\mathcal{A}} \rpa = \frac{\binom{v^{s-1}-1}{n_{\corr{x}_1}} \prod_{\tilde{x}_1\in\mathcal{A}\setminus \corr{x}_1} \binom{v^{s-1}}{n_{\tilde{x}_1}}}{\binom{v^s-1}{mv-1}},
\eeq
which gives averages
\beq
    \langle n_{\corr{x}_1} \rangle = \frac{mv-1}{v^s-1} \lpa v^{s-1} - 1\rpa = f \lpa v^{s-1} - 1\rpa,
\eeq
\beq
    \langle n_{\tilde{x}_1 \neq \corr{x}_1} \rangle = \frac{mv-1}{v^s-1} v^{s-1} = f v^{s-1}.
\eeq

Averaging the downward messages over the choices of rules $\rr$, we obtain:

\begin{itemize}
    \item for $x_1\neq \corr{x}_1$:
    \begin{align}
    \begin{split}
    \langle \tilde{\nu}_{\downarrow}(x_1) \rangle_\psi &= 
    \sum_{y}
    \nu_{\downarrow}(y)
    \langle
    \sum_{\substack{x_2,\dots,x_s\\
    \x\in\samp}}
    \langle \psi(y, \x) \rangle_{\asso}\ 
    \prod_{i=2}^s \nu_{\uparrow}(x_i)
    \rangle_{\samp}
    =\\
    & = \frac{m - \nu_{\downarrow}(\corr{y})}{mv-1} 
    \langle \delta_{\x\in\samp}\rangle_{\samp}
    \prod_{i=2}^s \sum_{x_i} \nu_{\uparrow}(x_i)
    = \frac{m - \nu_{\downarrow}(\corr{y})}{mv-1} f =\\
    &= \frac{m - \nu_{\downarrow}(\corr{y})}{mv-1}f,
    \end{split}
    \end{align}

    where $\langle \delta_{\x\in\samp}\rangle_{\samp} = \frac{\langle n_{x_1 \neq \corr{x}_1} \rangle}{v^{s-1}}$;
    
    \item for $x_1 = \corr{x}_1$:
    \begin{align}
    \begin{split}
    \langle \tilde{\nu}_{\downarrow}(\corr{x}_1) \rangle_\psi &= 
    \nu_{\downarrow}(\corr{y})\prod_{i=2}^s \nu_{\uparrow}(\corr{x}_i) +
    \sum_{y}
    \nu_{\downarrow}(y)
    \langle
    \sum_{\substack{x_2,\dots,x_s\\
    \x\in\samp, \x\neq\corr{\x}
    }}
    \langle \psi(y, \x) \rangle_{\asso}\ 
    \prod_{i=2}^s \nu_{\uparrow}(x_i)
    \rangle_{\samp}
    =\\
    &= \nu_{\downarrow}(\corr{y})\prod_{i=2}^s \nu_{\uparrow}(\corr{x}_i) +
    \frac{m - \nu_{\downarrow}(\corr{y})}{mv-1}
    \langle \delta_{\x\in\samp}\rangle_{\samp}
    \sum_{\substack{x_2,\dots,x_s\\
    \x\in\samp, \x\neq\corr{\x}
    }} \prod_{i=2}^s \nu_{\uparrow}(x_i) = \\
    &= \nu_{\downarrow}(\corr{y})\prod_{i=2}^s \nu_{\uparrow}(\corr{x}_i) +
    \frac{m - \nu_{\downarrow}(\corr{y})}{mv-1}
    f
    \lsq 1 - \prod_{i=2}^s \nu_{\uparrow}(\corr{x}_i)\rsq
    \end{split}
    \end{align}

    where $\langle \delta_{\x\in\samp}\rangle_{\samp} = \frac{\langle n_{\corr{x}_1} \rangle}{v^{s-1}-1}$.
\end{itemize}

The normalization factor is
\beq
    \langle Z_{\downarrow} \rangle_\rr = \sum_{x_1} \langle \tilde{\nu}_{\downarrow}(x_1) \rangle_\psi = 
    \nu_{\downarrow}(\corr{y})\prod_{i=2}^s \nu_{\uparrow}(\corr{x}_i) 
    + f\ \frac{m - \nu_{\downarrow}(\corr{y})}{mv-1}
    \lsq 1 - \prod_{i=2}^s \nu_{\uparrow}(\corr{x}_i)\rsq
    + (v-1) f \frac{m-\nu_{\downarrow}(\corr{y})}{mv-1}
\eeq
which gives the normalized average messages:
\begin{itemize}
    \item for $x_1 = \corr{x}_1$
\beq
    \langle \nu_{\downarrow}(\corr{x}_1) \rangle_\psi  = 
    \frac{
    \nu_{\downarrow}(\corr{y})\prod_{i=2}^s \nu_{\uparrow}(\corr{x}_i)
    + f\ \frac{m - \nu_{\downarrow}(\corr{y})}{mv-1} \lsq 1 - \prod_{i=2}^s \nu_{\uparrow}(\corr{x}_i)\rsq}{\langle Z_{\downarrow} \rangle_\rr};
    \label{eq:ave_nu_d}
\eeq
\item for $x_1 \neq \corr{x}_1$
\beq
    \langle \nu_{\downarrow}(x_1) \rangle_\psi  = 
    f \frac{
    \frac{m - \nu_{\downarrow}(\corr{y})}{mv-1}}{\langle Z_{\downarrow} \rangle_\rr}.
\eeq
\end{itemize}

\paragraph{Iterating over layers}
As for the upward process, the average messages for the downward process are of two kinds, one for the correct value $\corr{x}_1$ and one for the other values $x_1\neq\corr{x}_1$. To obtain a mean-field description of the BP process, we combine the average downward messages with the average upward ones by substituting $\nu_{\uparrow}(\corr{x}_i)\rightarrow \langle \nu_{\uparrow}(\corr{x}_i) \rangle$ in \autoref{eq:ave_nu_d}.
We use the notation 
\begin{align}
\langle \nnu{\ell}(\corr{x}_i) \rangle &= p_{\uparrow}^{(\ell)},\\
\langle \nnd{\ell}(\corr{x}_i) \rangle &= p_{\downarrow}^{(\ell)},
\end{align}
where the upwards and downwards beliefs $p_{\uparrow}^{(\ell)}$, $p_{\downarrow}^{(\ell)}$ in the correct value for the latent variable $X^{(\ell)}_i$ depend only on the layer $\ell$ and not on the specific position $i$ inside the layer.
Putting together \autoref{eq:iterUP} and \autoref{eq:ave_nu_d}, we obtain
\begin{align}
\begin{split}
    &p_{\uparrow}^{(\ell+1)} = F_{\uparrow}\lpa p_{\uparrow}^{(\ell)}\rpa,\\
    &p_{\downarrow}^{(\ell)} = F_{\downarrow}\lpa p_{\downarrow}^{(\ell+1)},  p_{\uparrow}^{(\ell)}\rpa,
\end{split}
\label{eq:anneal_iteration}
\end{align}
with
\begin{align}
    &F_{\uparrow}(p) = \frac{p^s + f \frac{m-1}{mv-1}\lpa 1-p^s\rpa}{p^s + f\lpa 1-p^s\rpa},\\
    &F_{\downarrow}(q, p) = \frac{q\ p^{s-1} + f \frac{m-q}{mv-1}\lpa 1-p^{s-1}\rpa}{q\ p^{s-1} + f \frac{m-q}{mv-1}\lpa 1-p^{s-1}\rpa + (v-1)f \frac{m-q}{mv-1}},
\end{align}

and the initialization condition
\begin{align}
    &p_{\uparrow}^{(0)} = 1-\epsilon + \epsilon/v,\\
    &p_{\downarrow}^{(L)} = 1/v.
\end{align}

From $p_{\uparrow}^{(\ell)}$ and $p_{\downarrow}^{(\ell)}$, at layer $\ell$, the average marginal probability of the correct value $p^{(\ell)}$ is given by
\beq
    p^{(\ell)} = \frac{p_{\uparrow}^{(\ell)} p_{\downarrow}^{(\ell)}}{p_{\uparrow}^{(\ell)} p_{\downarrow}^{(\ell)} + \frac{(1-p_{\uparrow}^{(\ell)})(1-p_{\downarrow}^{(\ell)})}{v-1}}.
    \label{eq:theory_marginal}
\eeq

\subsubsection{Validity of the mean-field theory}

\begin{figure}[H]
    \centering
    \includegraphics[width=\textwidth]{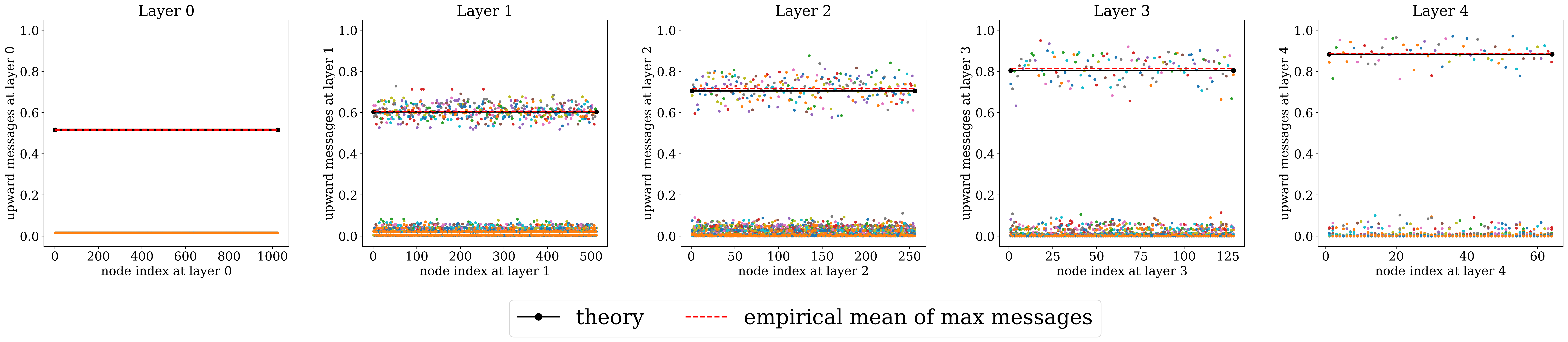}
    \vspace{-0.5cm}
    \caption{\textbf{Upward BP messages for layers $0$ to $4$ for $\epsilon=0.5$, $v=32$, $s=2$, $L=10$, $sf=0.5$.} Each node has $v$ messages, one per possible feature. At the input layer (layer $0$), messages have value $1-\epsilon=0.5$ or $\epsilon/v=0.5/32$. Going upward, the values of the messages fluctuate, but they stay separated into two distinct groups: large messages (i.e., the most probable feature for each node) and small ones. The annealed mean-field computation (represented with a black line) captures the mean value of the large messages well (red dashed line).}
    \label{fig:up-mean_messages}
\end{figure}

\begin{figure}[H]
    \centering
    \includegraphics[width=\textwidth]{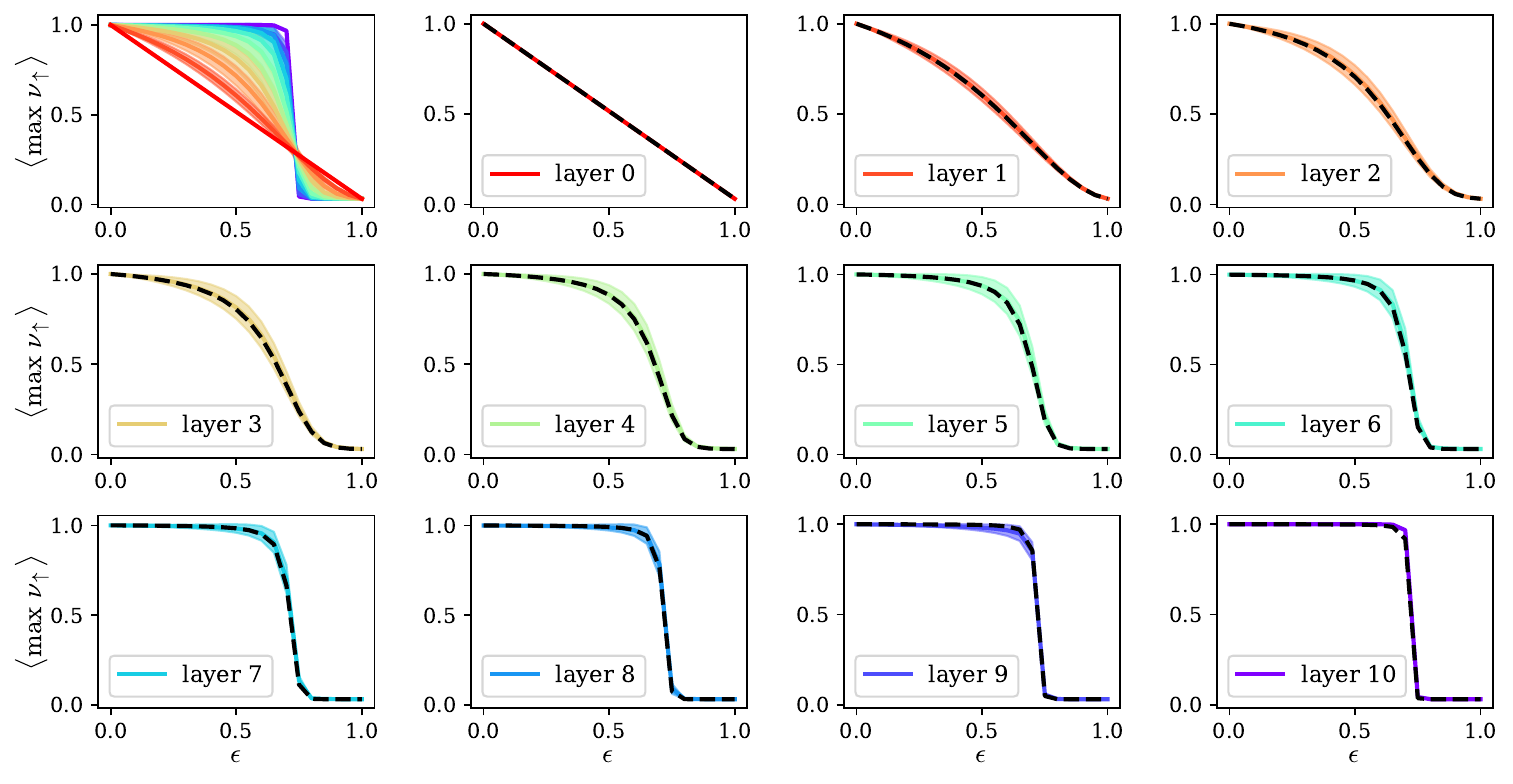}
    \vspace{-1cm}
    \caption{\textbf{Largest upward BP messages, averaged for each layer, for varying $\epsilon$. Data for the Random Hierarchical Model with $v=32$, $s=2$, $L=10$, $sf=0.5$.}
    Each layer, indicated in the legend, is represented with a different color, and the black dashed line is the theoretical prediction from \autoref{eq:anneal_iteration}, which shows excellent agreement with the experiments. The top left panel represents all the layers together for comparison. Starting from the initialization $\nu_{\uparrow}=1-\epsilon+\epsilon/v$ at layer $0$, we observe that the largest upward messages increase as we go to higher levels in the hierarchy only if $\epsilon$ is smaller than some threshold value. For $\epsilon$ larger than this threshold, instead, the messages become smaller, and it is not possible to reconstruct the highest levels in the hierarchy better than random chance.}
    \label{fig:upward_grid}
\end{figure}

\begin{figure}[H]
    \centering
    \includegraphics[width=\textwidth]{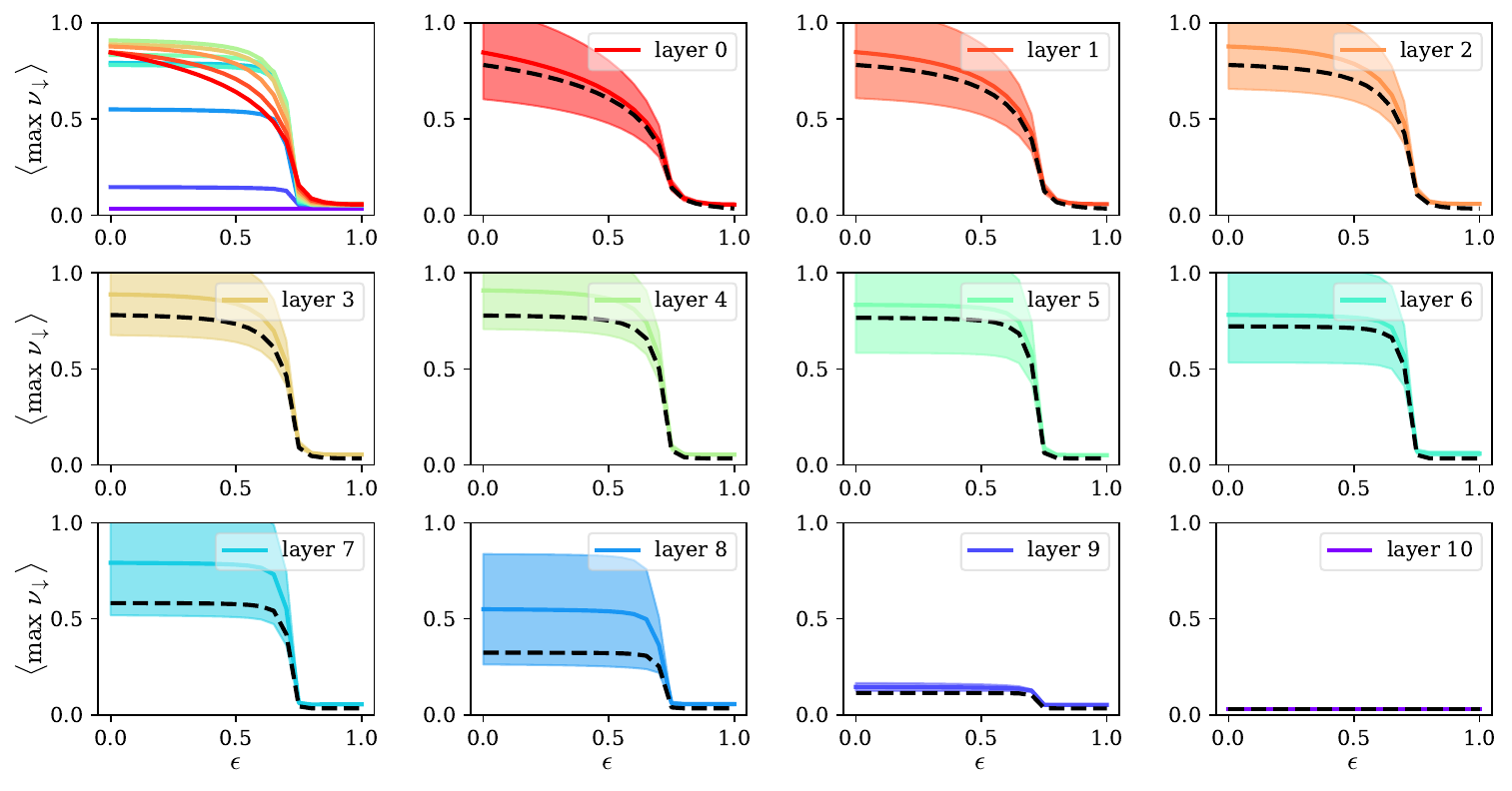}
    \vspace{-1cm}
    \caption{\textbf{Largest downward BP messages, averaged for each layer, for varying $\epsilon$. Data for the Random Hierarchical Model with the same parameters as \autoref{fig:upward_grid}.}
    Each layer, indicated in the legend, is represented with a different color, while the theoretical prediction from \autoref{eq:anneal_iteration} is represented with the black dashed line. We observe that the messages in the downward process have large fluctuations, as represented by their standard deviations, especially for small $\epsilon$. Still, the theory correctly captures the trend and becomes more accurate for increasing $\epsilon$. The top left panel represents all the layers together for comparison. Starting from the initialization $\nu_{\downarrow}=1/v$ at the top layer ($10$), we observe that the largest downward messages increase as we go to lower levels in the hierarchy only if $\epsilon$ is smaller than some threshold value.}
    \label{fig:downward_grid}
\end{figure}

Due to the randomness of the production rules, the messages $\nu_{\uparrow}(x)$, $\nu_{\downarrow}(x)$ are random variables that depend on the specific realization of the rules. Although their fluctuations are not captured by the averages computed in \autoref{eq:anneal_iteration}, we observe that $p_{\uparrow}^{(\ell)}$ and $p_{\downarrow}^{(\ell)}$ capture well the average behavior of the messages at a given layer. 
In \autoref{fig:up-mean_messages}, the values of $\nnu{\ell}(X_i^{(\ell)})$ are reported for the bottom $5$ levels of a Random Hierarchical Model with $L=10$, $s=2$, $v=32$, $m=8$, and noise level $\epsilon=0.5$. At each layer $\ell$, the index $i$ of the nodes goes from $1$ to $s^{L-\ell}$ and for each of them, there are $v=32$ messages $\nnu{\ell}$, one for each entry of the alphabet. At layer $\ell=0$, the messages $\nnu{0}$ are initialized according to \autoref{eq:belief}. After one iteration, at layer $\ell=1$, we observe that the largest messages at each node, those corresponding to the most probable features $x_i$, are spread around some mean value that is well captured by the theoretical prediction of \autoref{eq:anneal_iteration}. We observe that also for the upper layers, the average behavior of the largest messages is well captured by the theory.\\
The comparison between the theory and the BP algorithm for every $\epsilon$ is reported in the \autoref{fig:upward_grid} and \autoref{fig:downward_grid}. The upward iteration is reported in \autoref{fig:upward_grid} and shows an excellent agreement with the prediction of \autoref{eq:anneal_iteration}. In particular, going from the input layer $\ell=0$ to the class variable $\ell=L$ ($L=10$ in \autoref{fig:upward_grid}), the messages for the most probable features show a sharper transition at a threshold value $\epsilon^*$, which corresponds to the phase transition of the theoretical iteration map in \autoref{eq:iterUP} when $L\rightarrow\infty$.
The downward iteration in \autoref{fig:downward_grid} shows that the theory also captures the trend in $\epsilon$ of the downward messages. However, for small values of $\epsilon$, we observe that the messages have large fluctuations around their mean value. The reason for this behavior is that, in the Random Hierarchical Model, there is a number $m$ of synonyms $(x_1, \dots, x_s)$ that code for the same higher level feature $y$. Therefore, having perfect information on $y$ and on $x_2,\dots, x_s$ is not enough to perfectly reconstruct the value of $x_1$, thereby resulting in large fluctuations of the messages at small noise level $\epsilon$.
This is different from the upward process where having perfect information on $(x_1,\dots, x_s)$ allows the perfect reconstruction of $y$. As a result, the messages in the upward process are more concentrated around their mean than the downward messages.

\begin{figure}[H]
    \centering
    \includegraphics[width=\textwidth]{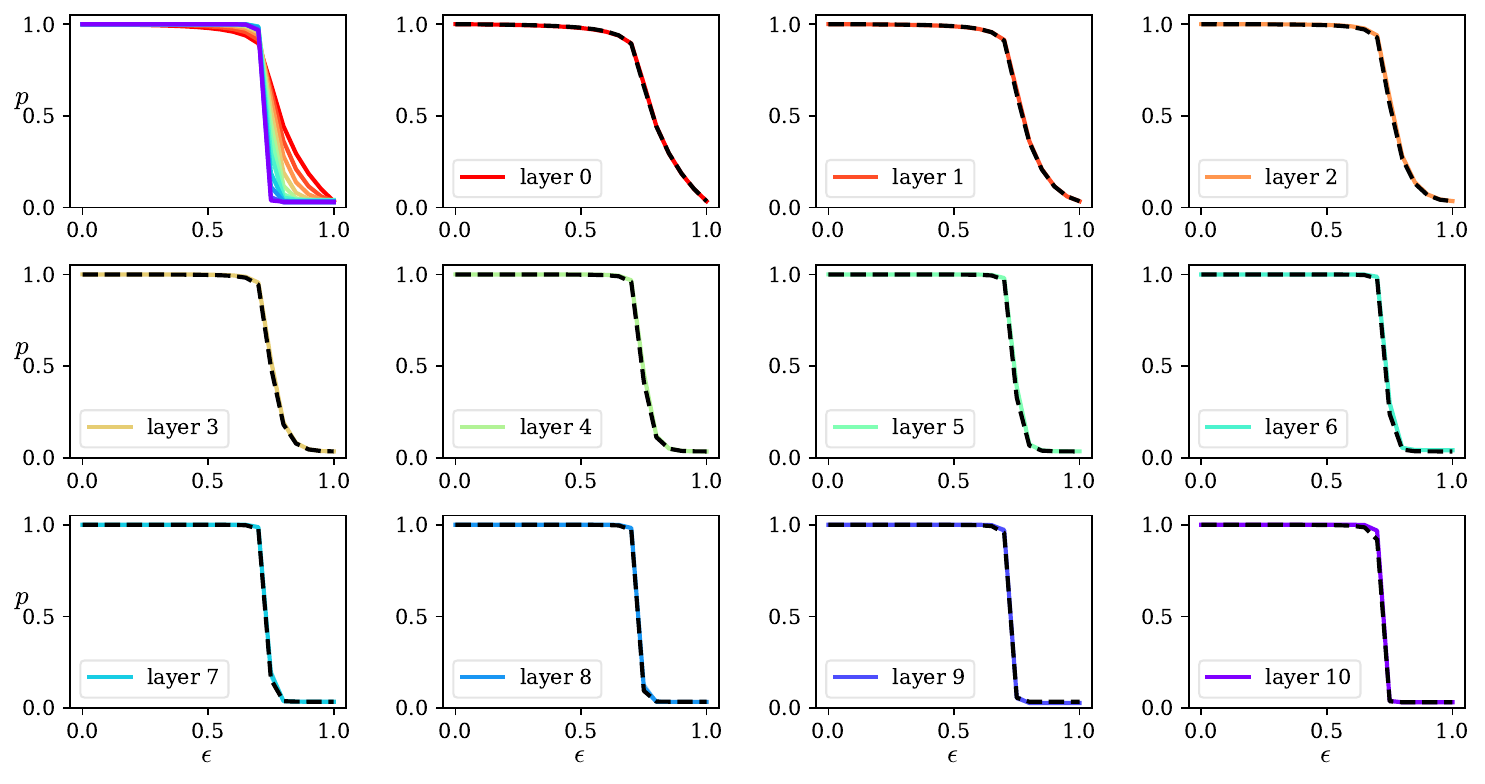}
    \vspace{-1cm}
    \caption{\textbf{Largest marginal probabilities computed by BP, averaged for each layer, for varying $\epsilon$. Data for the Random Hierarchical Model with the same parameters as \autoref{fig:upward_grid}.}
    Each layer, indicated in the legend, is represented with a different color, and the black dashed line is the theoretical prediction from \autoref{eq:theory_marginal}, which shows excellent agreement with the experiments. The top left panel represents all the layers together for comparison, where the inversion between the top and bottom layers can be observed (same curves as \figUDTheory in the main text).
}
    \label{fig:p_grid}
\end{figure}

\newpage

\hypertarget{app:CC}{\section{Mapping from time diffusion to $\epsilon$ noise}}
\label{app:CC}

\begin{figure}[H]
    \centering
    \includegraphics[width=.45\textwidth]{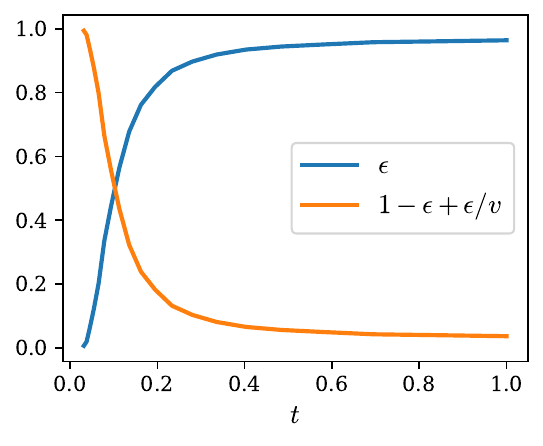}
    \includegraphics[width=.49\textwidth]{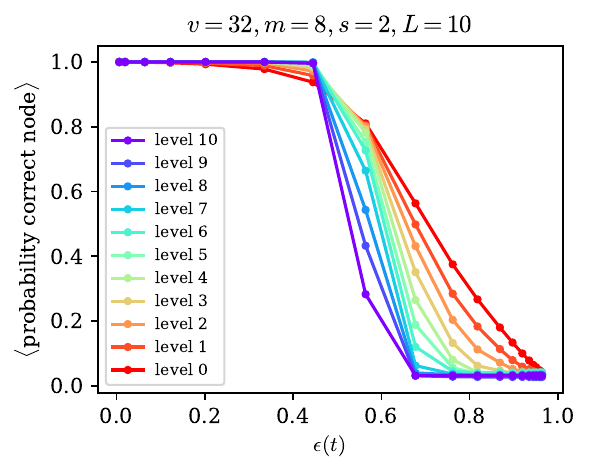}
    \caption{\textit{(Left panel)} \textbf{Mapping between the $\epsilon$ values and the diffusion process.} From the average values of the beliefs at the leaves variables during the diffusion process at time $t$, we compute an effective $\epsilon(t)$ as $\epsilon(t)/v = \langle \nnu{0}(x)\rangle$, for the values $x$ different from the starting one, averaging over the realizations of the diffusion process.
    \textit{(Right panel)} \textbf{Probability of reconstructing the initial values for the nodes at a given layer during the diffusion process in time, using the effective $\epsilon$ computed in the left panel.} We observe that the shape of the curves with respect to the effective noise $\epsilon(t)$ is qualitatively similar to that of the simplified $\epsilon$-process reported in \figUDTheory, supporting that it represents a good approximation for studying the diffusion process in time.}
    \label{fig:eps_time}
\end{figure}

In the diffusion process for the Random Hierarchy Model defined in \secBP, the beliefs $\nnu{0}$ at the input variables vary stochastically in time, according to \eqBayes. 
Instead, in the simplified model of noise considered in \secMF, at a given noise level $\epsilon$, these beliefs are fixed to two possible values (cf. \eqBeliefMain). To study whether the $\epsilon$-process is an effective approximation of the time diffusion process, we define an effective $\epsilon(t)$ depending on the reverse time of diffusion. At each input node $X^{(0)}_i$, we consider the upward messages $\nnu{0}(x)$ associated to the values $x$ that are different from the value of $X^{(0)}_i$ at time $t=0$. Denoting them as $\nu_t$, we define
\beq
    \frac{\epsilon(t)}{v} = \langle \nu_t \rangle,
\eeq
where the average $\langle \nu_t \rangle$ is performed over all the leaves variables $i$ and the realizations of the dynamics. $\epsilon(t)$ increases exponentially in time, according to the noise schedule used in the diffusion process, as shown in the left panel of \autoref{fig:eps_time}. The probability of correct reconstruction of a given node in the diffusion process is reported as a function of $\epsilon(t)$ in the right panel of \autoref{fig:eps_time}. We observe that the curves for different layers have similar behavior to those of the $\epsilon$-process presented in \figUDTheory of the main text, confirming that the latter is an effective approximation of denoising diffusion.

\hypertarget{app:resnets}{\section{Hidden activations for different architectures}}
\label{app:resnets}

\begin{figure}[H]
    \centering
    \begin{tikzpicture}
        \node[anchor=north west,inner sep=0pt] at (0,0){
        \includegraphics[width=0.3\textwidth]{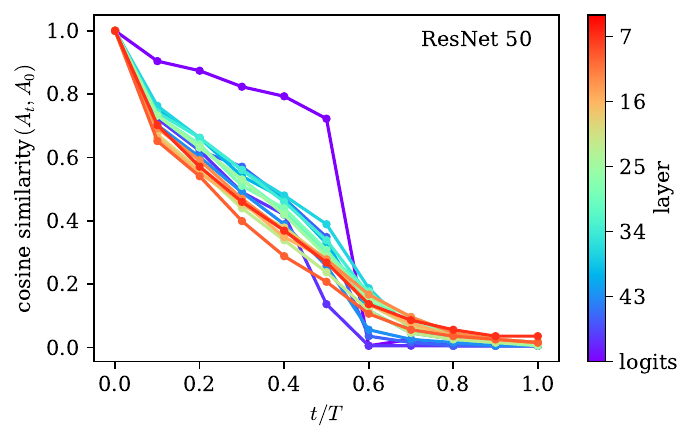}};
        \node[font=\large] at (0ex,-1ex) {(a)};
    \end{tikzpicture}
    \begin{tikzpicture}
        \node[anchor=north west,inner sep=0pt] at (0,0){
        \includegraphics[width=0.3\textwidth]{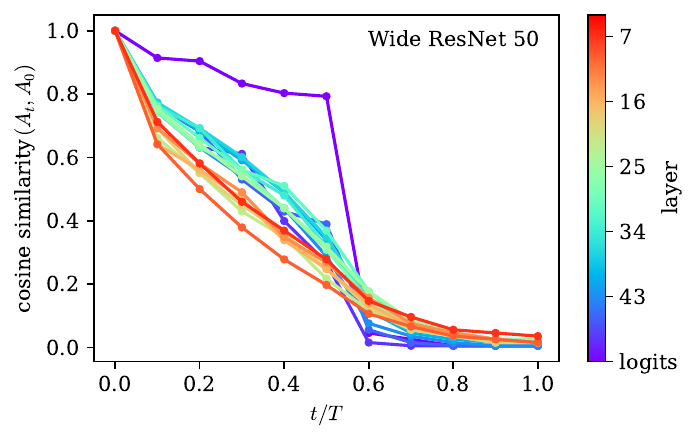}};
        \node[font=\large] at (0ex,-1ex) {(b)};
    \end{tikzpicture}
    \begin{tikzpicture}
        \node[anchor=north west,inner sep=0pt] at (0,0){
        \includegraphics[width=0.3\textwidth]{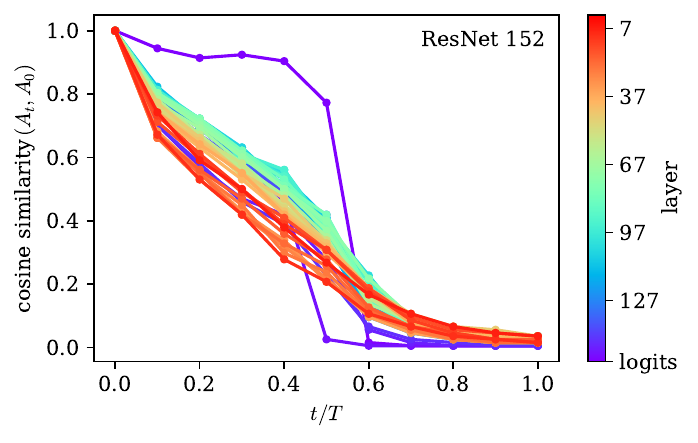}};
        \node[font=\large] at (0ex,-1ex) {(c)};
    \end{tikzpicture}
    \caption{\textbf{Cosine similarity between the post-activations of the convolutional blocks of different ResNet architectures for the initial images $x_0$ and the synthesized ones $\hat{x}_{0}(t)$.} Specifically, panels (a), (b), and (c) correspond to ResNet50, Wide ResNet50, and ResNet152, respectively \cite{he_deep_2016}. As in \figImageNet for the ConvNeXt architecture, the similarity between logits exhibits a sharp drop around $t \approx T/2$, indicating the change in class, while the hidden representations of the early layers change more smoothly.
    For computing the cosine similarity, all activations are standardized, i.e., centered around the mean and scaled by the standard deviation computed on the 50000 images of the ImageNet-1k validation set. At each time, the cosine similarity values correspond to the maximum of their empirical distribution over $10000$ images ($10$ per class of ImageNet-1k).
    }
    \label{fig:resnet}
\end{figure}

We perform the experiments described in {\secImageNetCNN} using the internal representations of different deep convolutional architectures trained for image classification on ImageNet-1k.
We consider the ResNet architecture \cite{he_deep_2016} with varying width and depth: a ResNet 50 achieving 95.4\% top-5 accuracy, a Wide ResNet 50 having 	
95.8\% top-5 accuracy, and a ResNet 152 having 96.0\% top-5 accuracy \cite{torchvision2016}.
The results of the experiments performed with the hidden representations of these architectures are reported in \autoref{fig:resnet} and show the same qualitative behavior as the one observed for the ConvNeXt architecture in \figImageNet: the cosine similarity exhibits a sharp transition for the logits, while it decays smoothly for the hidden representations at early layers.

\hypertarget{app:gaussian-mixture}{\section{Bi-modal distributions}}
\label{app:gaussian-mixture}

In this section, we study the forward-backward experiments discussed in the main text, focusing on a bi-modal distribution without hierarchical and compositional structure. Specifically, we consider a $d$-dimensional Gaussian mixture with an initial probability density given by:
\beq \label{eq:gauss-mixt}
    q(x_0) = \frac{1}{2(2 \pi \sigma^2)^{d/2}} \left[ \exp\left( - \frac{(x_0-\mu)^\top(x_0-\mu)}{2\sigma^2}\right) + \exp\left( - \frac{(x_0+\mu)^\top(x_0+\mu)}{2\sigma^2}\right)\right].
\eeq
We diffuse the data according to the dynamics described in Section 1.A of the main text, i.e.,
\beq
    x_t = \sqrt{1-\beta_t} x_{t-1} + \sqrt{\beta_t} \eta, \quad \eta \sim \mathcal{N}(0,\mathbb{I}).
\eeq
Thus, the \textit{forward dynamics} reads
\beq
    x_t = \sqrt{\overline{\alpha_t}} x_0 + \sqrt{1-\overline{\alpha_t}} \eta, \quad \eta \sim \mathcal{N}(0,\mathbb{I}).
\eeq
We then reverse the process at time $t$, following the exact \textit{backward dynamics}:
\beq
    x_{t-1} = \frac{1}{\sqrt{\overline{\alpha_t}}} \left( x_t + \beta_t \nabla_x \log q(x_t) \right) + \sqrt{\beta_t} z, \quad z \sim \mathcal{N}(0,\mathbb{I}),
\eeq
with the analytical \textit{score} function
\beq
    \nabla_x \log q(x_t) = -\frac{x_t}{\overline{\alpha_t} \sigma^2 + 1 -\overline{\alpha_t}} + \frac{\mu \sqrt{\overline{\alpha_t}}}{\overline{\alpha_t} \sigma^2 + 1 - \overline{\alpha_t}} \tanh \left( \frac{x_t^\top \mu \sqrt{\overline{\alpha_t}}}{\overline{\alpha_t} \sigma^2 + 1 - \overline{\alpha_t}} \right).
\eeq

As in our experiments on the ConvNeXt in Section 1 and on the deep CNN trained on the RHM in Section 4 of the main text, we examine how the internal representations of a neural network trained to classify the mode of the distribution vary when the input is obtained by inverting the forward dynamics at time $t$. 

We consider the Gaussian mixture defined in \autoref{eq:gauss-mixt} with $d=1024$, $\mu=(1,1,\dots,1)^\top$, and $\sigma=1$. We train a deep, fully-connected ReLU network with 6 hidden layers, each containing 64 neurons, using $n=2048$ training points until achieving zero training error. This network achieves $100\%$ accuracy on a test set with $n_{\rm test} = 1024$ samples. 

\begin{figure}[H]
    \centering
    \includegraphics[width=0.45\linewidth]{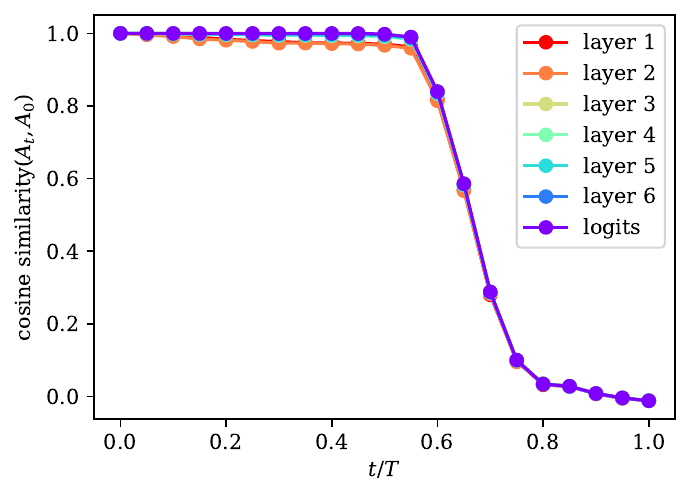}
    \caption{\textbf{Cosine similarity between the post-activations of the hidden layers of a deep fully-connected network for bi-modal data} $x_0 \in \mathbb{R}^d$ ($d=1024$) and the ones synthesized with forward-backward experiments $\hat{x}_0(t)$. Around $t \approx 0.6\,T$, the similarity between logits exhibits a drop, indicating a transition in the probability of changing the initial mode. In contrast to the RHM and natural images, the hidden representations of the hidden layers change like the logits. In particular, no crossing of the curves is observed. To compute the cosine similarity, all activations are standardized, i.e., centered around the mean and scaled by the standard deviation computed on 1000 initial samples.}
    \label{fig:gaussian}
\end{figure}

For each inversion time $t$, we compute the cosine similarity between the post-activations for the initial and generated points. \autoref{fig:gaussian} presents the resulting curves. Similar to the curves obtained for image and synthetic hierarchical data, the class similarity curve exhibits a drop at a characteristic time, as theoretically studied by Biroli et al. in \cite{biroli2023generative} and \cite{biroli2024dynamical}. However, unlike compositional data, the behavior of the curves corresponding to the internal layers follows the curve for the class. Specifically, there is no inversion of the similarity curves corresponding to early and deep layers, which is a phenomenon unique to compositional data (\figImageNetAct and \figUDAct) and cannot be captured by simple bi-modal distributions.

\end{document}